\documentclass[manuscript,screen,review=false,acmsmall]{acmart}
\usepackage{hyperref}
\usepackage{multirow}
\usepackage{color, colortbl}

\usepackage{cellspace}
\definecolor{Gray}{gray}{0.9}

\definecolor{arsenic}{rgb}{0.23, 0.27, 0.29}
\definecolor{airforceblue}{rgb}{0.36, 0.54, 0.66}

\usepackage{soul}

\usepackage{framed}

\AtBeginDocument{%

  \providecommand\BibTeX{{%
    \normalfont B\kern-0.5em{\scshape i\kern-0.25em b}\kern-0.8em\TeX}}}

\setcopyright{acmcopyright}
\copyrightyear{2024}
\acmYear{2024}
\acmDOI{XXXXXXX.XXXXXXX}

\acmConference[Accepted at CSCW'24]{Make sure to enter the correct
  conference title from your rights confirmation emai}{2024}{San José, Costa Rica}
\acmPrice{15.00}
\acmISBN{978-1-4503-XXXX-X/18/06}

\begin{document}

\title[A Framework for Mitigating the Use of Protected Attributes]{NLPGuard: A Framework for Mitigating the Use of Protected Attributes by NLP Classifiers}

\renewcommand{\shortauthors}{Greco et al.}

\author{Salvatore Greco}
\authornote{Work done at Nokia Bell Labs.}
\email{salvatore\_greco@polito.it}
\orcid{0000-0001-7239-9602}
\affiliation{%
  \institution{Politecnico di Torino}
  \city{Turin}
  \country{Italy}
}

\author{Ke Zhou}
\email{ke.zhou@nokia-bell-labs.com}
\orcid{0000-0001-7177-9152}
\affiliation{%
  \institution{Nokia Bell Labs}
  \city{Cambridge}
  \country{UK}}

\author{Licia Capra}
\email{l.capra@ucl.ac.uk}
\orcid{0000-0003-1425-3837}
\affiliation{%
  \institution{University College London}
  \city{London}
  \country{UK}
}

\author{Tania Cerquitelli}
\email{tania.cerquitelli@polito.it}
\orcid{0000-0002-9039-6226}
\affiliation{%
 \institution{Politecnico di Torino}
 \city{Turin}
 \country{Italy}}

\author{Daniele Quercia}
\email{daniele.quercia@nokia-bell-labs.com}
\orcid{0000-0001-9461-5804}
\affiliation{%
  \institution{Nokia Bell Labs}
  \city{Cambridge}
  \country{UK}}



\begin{abstract}

AI regulations are expected to prohibit machine learning models from using sensitive attributes during training. However, the latest Natural Language Processing (NLP) classifiers, which rely on deep learning, operate as black-box systems, complicating the detection and remediation of such misuse. Traditional bias mitigation methods in NLP aim for comparable performance across different groups based on attributes like gender or race but fail to address the underlying issue of reliance on protected attributes. To partly fix that, we introduce \textsc{NLPGuard}, a framework for mitigating the reliance on protected attributes in NLP classifiers. \textsc{NLPGuard} takes an unlabeled dataset, an existing NLP classifier, and its training data as input, producing a modified training dataset that significantly reduces dependence on protected attributes without compromising accuracy. \textsc{NLPGuard} is applied to three classification tasks: identifying toxic language, sentiment analysis, and occupation classification. Our evaluation shows that current NLP classifiers heavily depend on protected attributes, with up to 23\% of the most predictive words associated with these attributes. However, \textsc{NLPGuard} effectively reduces this reliance by up to 79\%, while slightly improving accuracy. \break

\noindent \textit{\textbf{\underline{Disclaimer}: This paper contains examples of language that some people may find offensive.}}

\end{abstract}

\keywords{protected attributes, bias, fairness, natural language processing, toxic language, large language models, crowdsourcing}




\maketitle

\renewcommand{\shortauthors}{Greco et al.}

\section{Introduction}
\label{intro}
In recent years, the adoption of deep learning-based NLP models has exponentially increased. Transformer-based models, such as BERT \cite{devlin-etal-2019-bert}, T5 \cite{DBLP:journals/corr/abs-1910-10683}, and GPT \cite{DBLP:journals/corr/abs-2005-14165}, have achieved unthinkable levels of performance on several natural language tasks. However, despite being increasingly accurate, these models remain black-boxes \cite{10.1145/3236009}. For an NLP classification task, models predict a class label from an input text without providing any information on the complex internal decision-making mechanism, making it challenging to identify and mitigate potential bias and/or unfair behavior in such models.

Upcoming privacy laws regulating the use of AI will soon demand that learning shall not be done on protected attributes such as race, gender, or sexual orientation, as already identified by the \textit{General Data Protection Regulation} (GDPR), the \textit{UK Government}, and the anti-discrimination legislation in the United States~\cite{chouldechova2020snapshot,Goodman_Flaxman_2017,Floridi2022}. Ensuring AI models avoid using protected attributes in decision-making is termed `{\em fairness through unawareness}' \cite{10.1145/3457607}, and it is crucial in many real-world scenarios. For instance, NLP-based systems often assess job applicants' resumes. Following the Civil Rights Act in the US, discrimination based on race, sex, nationality, or other protected attributes is forbidden. Hence, these NLP systems must omit words linked to protected attributes to prevent discriminatory practices against candidates, such as the ``sexist'' Amazon Recruitment tool,\footnote{\url{https://www.bbc.com/news/technology-45809919}} a system that learned to downgrade resumes containing the word \textit{`women'}. Content moderation is another example, where all users should be treated equitably, without having their contributions censored or suppressed because of, for example, their demographic characteristics.

\begin{figure}[]
    \centering
   
    \begin{minipage}{0.50\textwidth}
        \centering

\scriptsize
\begin{tabular}{cc}
\toprule
\textbf{Text}                                                            &  \textit{\textbf{P(T)}} \\ \toprule
\textit{"I like this city! There are many black people!"}                          & 0.53                   \\ \midrule
\textit{"The homosexual marriage bill will be debated soon! I am in favor!"}           & 0.62                   \\ \midrule
\textit{"This city is incredibly modern! If you are gay, you are not judged."}          & 0.88                  \\ \midrule
\textit{"I hate this fucking shitty city! There are many black people!"}           & 0.99                   \\ \bottomrule
\end{tabular}
\vspace{-3.6mm}
\caption{\textbf{Toxicity probabilities P(T) to four sentences predicted as toxic by a classifier}. The first three sentences are misclassified, while the last is correctly identified.}
\label{tab:motivation-example-original-model}
    \end{minipage}
    \hfill
    \begin{minipage}{0.44\textwidth}
        \centering
  \includegraphics[clip,width=0.98\columnwidth]{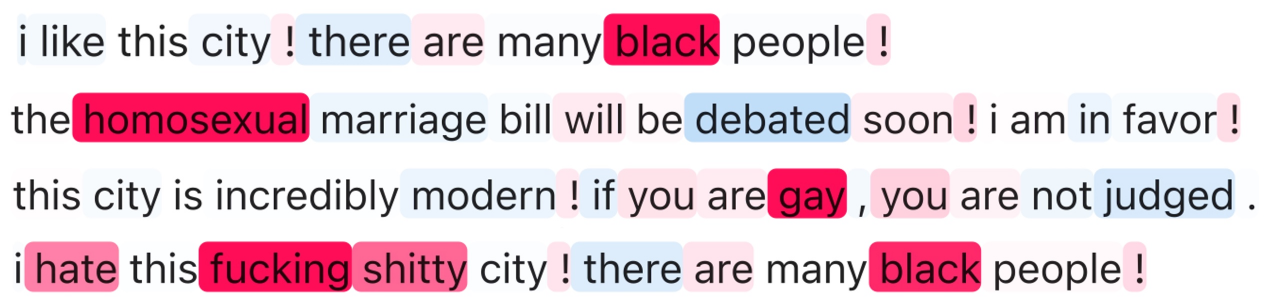}%
\vspace{-3mm}
\caption{\textbf{Words impacting the toxicity classification of the four sentences in Figure~\ref{tab:motivation-example-original-model}}. The more intense the red (blue) color of a word, the more important the word contributes to toxic (non-toxic) classification. }
\label{fig:motivation-example-original-model}

    \end{minipage}
\end{figure}

However, as we will demonstrate in our analysis, state-of-the-art models often base their predictions on protected attributes, and accurate ones are frequently black boxes, posing challenges in identifying such misuse. Consider, for example, the task of determining whether a sentence contains toxic language or not in a dataset we will analyze. In Figure~\ref{tab:motivation-example-original-model}, we report four example sentences, together with the outcome of a toxicity classifier P(T); in Figure~\ref{fig:motivation-example-original-model}, we highlight in red the important words used by the classifier to make these predictions. As shown, the presence of words such as \textit{`black'}, \textit{`gay'}, or \textit{`homosexual'} is used to distinguish between toxic or non-toxic texts. Yet, these words are protected attributes and should not be used in such classifications at all. 

As discussed in \S\ref{sec:related-works}, prior studies on bias in NLP primarily focused on two challenges: ensuring fair performance across different groups and rectifying unfairness in word representations. However, these solutions only target specific biases and fail to eliminate the reliance of models on protected attributes for predictions. Therefore, we propose methods to reduce this bias in black-box NLP classifiers, removing most protected attributes from their decision-making process while maintaining accuracy, and making these approaches applicable across various datasets and tasks. 

In so doing, we make four main contributions:

\begin{enumerate}
    \item We introduce \textsc{NLPGuard} (\S\ref{sec:sec-3-our-framework}), a framework with three components: (1) an {\em Explainer} that finds the most important words for predictions; (2) an {\em Identifier} that checks if these words are about protected attributes; and (3) a {\em Moderator} that adjusts the training data to re-train the NLP model to reduce learning from such protected attributes.

    \item We evaluate each part of our framework and use it to mitigate toxicity detection in Wikipedia comments with BERT (\S\ref{sec:results-performance}). BERT depends on protected attributes for toxicity predictions (23\% of the most predictive words), but our approach cuts this down by 60\% and even increases prediction accuracy by 0.8\%.

    \item We then evaluate whether our framework generalizes to different types of data and tasks, not just toxicity detection (\S\ref{sec:results-general}).
    We found that our framework reduces the use of protected attributes by 79\% when applied to out-of-distribution data. Also, it reduced reliance on protected attributes without compromising accuracy in tasks like sentiment analysis and occupation classification.

    \item We make \textsc{NLPGuard} publicly accessible,\footnote{The code repository of our framework is available at \url{https://github.com/grecosalvatore/nlpguard}} and discuss how to incorporate it into existing NLP systems, its impact, and its limitations (\S\ref{sec:discussion}).

\end{enumerate}


\section{Related Work}
\label{sec:related-works}

\subsection{AI Regulations and Laws}
\label{ssec:ai-regulations}
\noindent
The growth of AI systems has raised privacy and discrimination concerns, leading to the introduction of numerous regulations and laws governing their use.
In the European Union (EU), in May 2018, the GDPR \cite{gdpr} was introduced, which demands organizations ensure that personal data is processed lawfully, fairly, and transparently. It prohibits processing sensitive personal attributes such as race, ethnicity, religion, and political opinions, unless legitimately justified. The EU proposed the AI Act \cite{theaiact,theaiactdoc}, which defines rules and obligations depending on the level of risk of AI systems (e.g., transparency, documentation, human oversight) \cite{10.1145/3593013.3594069}. 
In the United Kingdom (UK), the UK Equality Act 2010 \cite{ukequalityact} established that it is unlawful to discriminate based on nine protected characteristics: age, disability, gender reassignment, marriage and civil partnership, pregnancy and maternity, race, religion or belief, sex, and sexual orientation. Compliance with the act is enforced by the Equality and Human Rights Commission (EHRC). In the United States (US), the Anti-discrimination Act~\cite{usantidiscriminationact} safeguards individuals from unfair treatment based on protected attributes. In late 2022, a blueprint of the AI Bill of Rights was passed \cite{whitehouse}, declaring that algorithms that discriminate or perform unjustified different treatment based on protected attributes violate legal protections. AI regulations will continue to evolve in the coming years \cite{Jobin2019,doi:10.1177/2053951716679679}, driven by a common goal: to minimize discriminatory outputs based on protected characteristics~\cite{Floridi2022,chouldechova2020snapshot,Goodman_Flaxman_2017}. 

\subsection{Bias mitigation for NLP}
\label{ssec:bias-mitigation}

Bias in NLP decision-making has manifested itself in several ways, including dialogue generation~\cite{dinan2019build}, text classification~\cite{10.1145/3278721.3278729}, and machine translation~\cite{stanovsky-etal-2019-evaluating}. It usually arises from training data \cite{app11073184, sun2019mitigating}.  For instance, pre-trained models and word embedding can inherit biases and stereotypes present in the large training corpora \cite{DBLP:journals/corr/BolukbasiCZSK16a,doi:10.1126/science.aal4230,DBLP:journals/corr/abs-1903-03862,ravfogel-etal-2020-null}. When quantifying bias, existing works generally highlight disparities between demographic groups, with differences in performance or selection bias on protected attributes such as race, gender, religion, and sexual~orientation \cite{app11073184, sun2019mitigating, kostick2022mitigating,10.1145/3465416.3483299}.

To address biases in NLP, techniques can be developed that act at the three main stages of the NLP pipeline \cite{a00dbc5bf49b4e44b71a5391b37211d9,KOZODOI20221083}: \textit{pre-processing} (modifying training data), \textit{in-processing} (imposing fairness constraints during model training), and \textit{post-processing} (adjusting classifier predictions based on fairness metrics). Most existing works focus on the first two stages, exploiting data augmentation and modified model training techniques \cite{10.1145/3308558.3313504, 10.1145/3278721.3278729,park2018reducing,ravfogel-etal-2020-null,zhang2020demographics,attanasio-etal-2022-entropy,sen-etal-2022-counterfactually}.
Furthermore, most of those studies focus on one protected category at a time. For example, \citet{10.1145/3308558.3313504} proposed the identification of protected attributes, such as gender, by creating a manual list of words, measuring the skewed occurrence of words across classes or predicted class probability distribution of words. \citet{park2018reducing} introduced gender swapping to equalize the number of male and female entities in the training data. \citet{10.1145/3278721.3278729} proposed dataset augmentation strategies that generate new sentences using templates or replace protected attributes with generic tags, such as part-of-speech or named-entity tags. \citet{zhang2020demographics} proposed mitigating biases in the training data by assuming a non-discrimination distribution and then reconstructing the distribution using instance weighting. \citet{ravfogel-etal-2020-null} proposed removing information from neural representations concerning gender or race for debiasing word embedding for NLP classification.

These past works try to mitigate unintended bias and performance imbalance between subgroups by (1) removing implicit bias from word embeddings, (2) performing data augmentation on the training set (data-based), or (3) intervening directly in the model architecture or objective function (model-based). However, a gap remains in evaluating (and tackling) the extent to which NLP classifiers depend on protected attributes for their predictions. In this paper, we aim to fill that gap. We consider the following definition of {\em fairness through unawareness}: ``an algorithm is fair as long as any protected attributes are not explicitly used in the decision-making process'' \cite{10.1145/3457607,NIPS2017_a486cd07,GrgicHlaca2016ProcessFairness}. Textual data is unstructured; hence, protected attributes are not explicitly delineated as input features, such as columns used in structured datasets. Consequently, we refine this definition in the context of NLP applications to ensure words associated with protected characteristics are not utilized in decision-making unless necessary. Our approach aims to reduce the use of protected attributes in the decision-making process of NLP models, thereby better aligning them with legal regulations.

Compared to prior work, our approach not only has a different objective, but it also overcomes two of their main limitations: (1) their focus on a subset of protected attributes at a time (usually race and gender); (2) their manual and static identification of protected attributes via pre-defined dictionaries, lists of identity terms, or additional annotations. The only technique addressing these limitations is \textit{Entropy-based Attention Regulation} (EAR) \cite{attanasio-etal-2022-entropy}. EAR introduces a regularization term to discourage overfitting to training-specific potentially biased terms.  However, those terms are automatically identified during training, leaving no flexibility for users to select which categories to mitigate.  Unlike previous techniques, our approach: (1) identifies and mitigates multiple protected categories simultaneously; (2) can be fully automated, allowing for a dynamic update of the dictionary of protected attributes; and (3) allows for the selection of the categories to mitigate.

\section{Our Mitigation Framework}
\label{sec:sec-3-our-framework}
Our Mitigation Framework, namely \textsc{NLPGuard}, has been designed to be generally applicable to any supervised machine learning-based NLP classification model applied on an unlabelled corpus. As illustrated in Figure~\ref{fig:architecture}, \textsc{NLPGuard} takes in input an unlabeled corpus and a pre-trained NLP classifier (together with its training dataset) to produce a mitigated training dataset in output. Ground truth class labels for the unlabelled corpus are not required; rather, the classifier is used to generate them, both for in-distribution (i.e.,~data that comes from the same distribution as the original training dataset) and for out-of-distribution data where labels are unavailable. Because of the black-box nature of NLP classifiers based on deep learning models, labels might be predicted using protected attributes. To mitigate that, our framework comprises the following three components:

\begin{figure*}
    \centering
   \includegraphics[width=0.99\textwidth]{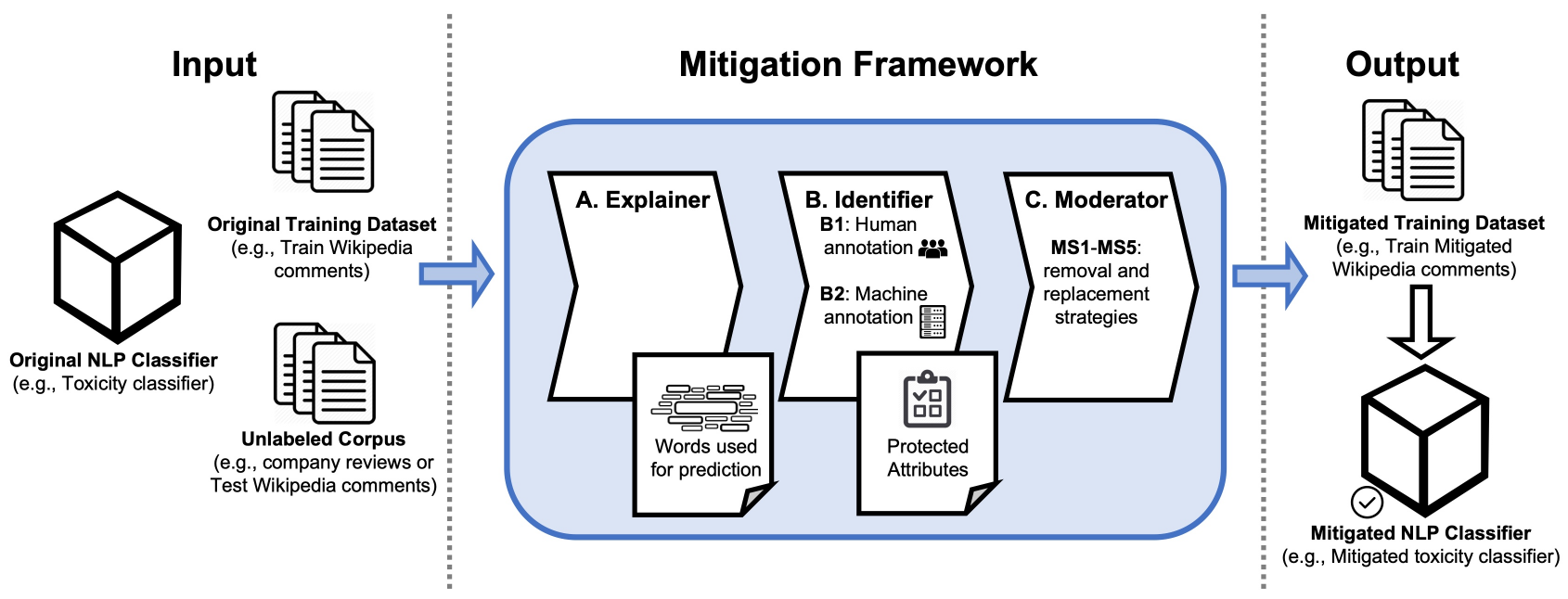}
\vspace{-5mm}
    \caption{\textbf{Our Mitigation Framework.} It takes the original NLP classifier, the original training dataset, and a new unlabeled corpus as input. The framework consists of three components: A) an \textit{Explainer} that identifies the most important words used by the classifier for predictions on the unlabeled corpus; B) an {\em Identifier} that determines which of those words are protected attributes; and C) a \textit{Moderator} that generates a mitigated training dataset to re-train the classifier so to reduce reliance on the previously identified protected attributes. }
    \label{fig:architecture}
\end{figure*}

\vspace{1mm}
\noindent
\textbf{A. Explainer.} This component uses \textit{Explainable Artificial Intelligence} (XAI) techniques to identify the most important words used by the model for its predictions. The XAI field has made great strides in making black-box models more transparent, and several techniques exist to explain NLP classifiers \cite{danilevsky-etal-2020-survey}. The best one for our purpose should have two qualities: (1) quantify the importance of each feature word (feature-based), and (2) be applicable to explain the model's predictions after training (post-hoc). Many techniques meet these requirements, and most of them measure the importance of each word for the prediction within an individual sentence (local-explanations) \cite{DBLP:journals/corr/RibeiroSG16,NIPS2017_7062,Ventura2022,8237336,arras-etal-2016-explaining,10.5555/3305890.3306024}. Our Explainer component first identifies the words important for prediction within all individual sentences, exploiting any of those techniques.  Each word is, as such, associated with multiple scores, one for each occurrence in each sentence. Second, it determines the most important predictive words for the model as a whole (global-explanations) following the idea of some of these techniques, which aggregate the words' importance over many sentences to compute the overall importance \cite{vanderlinden2019global,Ventura2022}. Specifically, for each word, it sums all their individual scores and divides them by their frequency to compute the word's classification score. The normalization step is required to also identify rare but important words. The output of the \textit{Explainer} component is the ordered list of the most important words for the model's predictions.

\vspace{1mm}
\noindent
\textbf{B. Identifier.} This component aims to annotate which of the previously detected important words refer to protected attributes. We consider the nine protected categories defined by the Equality Act: \textit{age}, \textit{disability}, \textit{gender reassignment}, \textit{marriage and civil partnership}, \textit{pregnancy and maternity}, \textit{race}, \textit{religion or belief}, \textit{sex}, and \textit{sexual orientation}. Our framework allows for annotation using \textit{human-in-the-loop} (B1) and \textit{machine-in-the-loop} (B2) approaches.  

\vspace{1mm}
\textbf{B1. Human-in-the-loop Annotation.} Crowdsourcing platforms, such as Amazon Mechanical Turk (MTurk) and Prolific \cite{10.1007/978-3-642-35142-6_14}, have been extensively used by the research community to recruit crowdworkers for data labeling purposes. Crowdworkers \cite{10.1145/1753846.1753873,10.1145/3492854} are anonymous people usually paid for completing simple tasks. This component leverages crowdsourcing to perform the protected attribute annotation of the most important words. In our work, we exploited MTurk, where for each important word \textcolor{airforceblue}{ \textsc{\{word\}}}, participants were asked to answer the following question:
\begin{description}
    \item \footnotesize  \textbf{Question:}  \textit{Is the word \textcolor{airforceblue}{\textsc{\{word\}}} referring to:} 
    \item \footnotesize \textbf{Possible Answers:} \footnotesize 1. \textit{Age}, 2. \textit{Disability}, 3. \textit{Gender reassignment}, 4. \textit{Marriage and civil partnership}, 5. \textit{Pregnancy and maternity}, 6. \textit{Race}, 7. \textit{Religion or belief}, 8. \textit{Sex}, 9. \textit{Sexual orientation}, 10. \textit{None of the above};
\end{description}
\noindent
To ensure data quality, we adopt a trap mechanism to detect random responses from participants and reject them (details of this mechanism are presented in \S \ref{ssec:protected-attribute-identifier-evaluation}).

\vspace{1mm}
\textbf{B2. Machine-in-the-loop Annotation.} As a cost-effective and scalable alternative to human-in-the-loop annotations via crowdsourcing, we also implemented a protected attributes annotation process that uses Large Language Models (LLMs). We did so inspired by a recent study \cite{gilardi2023chatgpt} that found LLMs, including ChatGPT, to outperform crowdworkers in text-based annotation tasks. It also has been shown that LLMs are effective in solving many NLP tasks \cite{qin2023chatgpt}.  Specifically, we implemented an annotation process that interacts with ChatGPT as follows: two prompts provide the protected categories, their definitions, and links with additional information. Then, for each word, the LLM is asked to: (1) classify the word into one of the protected categories or none of them; (2) provide a reliability score in the range $[0, 100]$; and (3) provide an explanation. For example, the word \textit{`homosexual'} would be classified with the protected category \textit{sexual orientation} and a score of $100/100$ by GPT-3.5-Turbo (see Figures \ref{fig:prompt1} -- \ref{fig:prompt4} in Appendix A for further details).

\vspace{1mm}
\noindent
\textbf{C. Moderator.} This component produces a new mitigated training dataset that can be used to train a new classifier that uses fewer protected attributes previously identified. It takes the original pre-trained classifier, the original training dataset, and the list of the most important words enriched with the protected attribute label as input. It produces a new \textit{mitigated training dataset} by adjusting the training dataset based on the identified protected attributes that can then be used to train a new mitigated classifier.  We designed and tested five mitigation strategies (MS).

\textbf{(MS1) Sentence-level removal.} Previous works have shown that subsampling can be an easy but effective technique for data balancing \cite{pmlr-v177-idrissi22a,10.5555/3524938.3525711}. This mitigation strategy eliminates all sentences containing protected attributes from the training set. As a result, it reduces the overall number of training examples. For example, if a word $W_i$ is identified as a protected attribute, all sentences in the training set that include that word are removed. The idea behind this strategy is that the imbalance in the number of training examples containing protected attributes for a particular class may have led the model to learn that these protected attributes are crucial for classifying that class.

\textbf{(MS2) Word-level removal.} This strategy removes only the protected attribute words from the sentences in the training set while preserving the number of examples. The process involves removing the identified protected attribute words from all sentences in the training set, thereby removing their influence on the model's learning process. The idea is that the model should be able to classify sentences without relying solely on the protected words, and rather use other words in the text too.

\textbf{(MS3) Word-level replacement with a random synonym.} This strategy replaces every instance of a protected attribute in the training set with one of its synonyms. It first uses embedding similarity techniques to identify the k-nearest neighbors for each protected attribute. Then, it randomly selects one of the k-most similar words to replace each instance of the protected attribute in the training set. This has been shown to mitigate bias in \cite{10.1145/3308558.3313504}, and it is believed that it may also help mitigate the use of protected attributes in classification, as the model may learn to rely on other words for classifying the classes rather than solely relying on protected attributes. This approach maintains the same number of examples in the training set but increases the diversity of words.

\textbf{(MS4) Word-level replacement with K random synonyms.} This strategy expands the training set by generating new sentences using synonyms of protected attributes. Instead of replacing the protected attribute in-place with one similar word as in \textit{MS3}, it creates \textit{k} new sentences by replacing the protected attribute with each of its k-nearest neighbors. For example, given a sentence containing a protected attribute $W_i$, \textit{k} new sentences are created by replacing $W_i$ with each of its k most similar words. This increases the size of the training set and diversifies the words used in the sentences.

\textbf{(MS5) Word-level replacement with hypernym.} This strategy replaces instances of protected attributes in the training set with higher-level words, called hypernyms, which provide a more general representation of the category to which the protected attribute belongs. For example, the hypernym of \textit{`dog'} could be \textit{`animal'}. By using hypernyms instead of the specific protected attributes, the model may not discriminate based on these attributes in its classifications. This technique has been shown to be effective in mitigating accuracy imbalance between subgroups~\cite{10.1145/3308558.3313504}.

\section{Framework evaluation: effectiveness and sensitivity}
\label{sec:results-performance}
We evaluate the effectiveness of our framework and the sensitivity of the \textit{Explainer} (\S \ref{ssec:explainer-evaluation}), \textit{Identifier} (\S \ref{ssec:protected-attribute-identifier-evaluation}), and \textit{Moderator} (\S \ref{ssec:moderator-evaluation}) components in mitigating a toxicity classifier applied to in-distribution data (i.e., the test set). 

\subsection{Evaluation task}
\label{ssec:in-distribution-task}
We choose toxicity prediction as the main evaluation task in line with previous research. Toxicity classifiers are used in different contexts \cite{kumar2021designing,10.1145/3415179}, such as Reddit, Twitter, and 4chan, with competitive performance. However, they suffer from different types of biases \cite{sap-etal-2019-risk,DBLP:journals/corr/DavidsonWMW17, DBLP:journals/corr/abs-1905-12516,10.1145/3531146.3533144}. In Wikipedia, for example, any comment containing words associated with insults, offense, or profanity, regardless of the tone, the intent, and the context, would be classified as toxic; toxic language was however more likely predicted from minority communities, as found in  \cite{DBLP:journals/corr/abs-1905-12516}, thus suggesting the use of protected attributes by such models.

In our experiments, we used the ``original model'' in the widely used detoxify \cite{Detoxify}
library{\footnote{\url{https://github.com/unitaryai/detoxify}}} as a pre-trained toxicity classifier. This is a BERT-base and uncased model \cite{devlin-etal-2019-bert} trained on a dataset of publicly available Wikipedia comments.\footnote{\url{https://www.kaggle.com/competitions/jigsaw-toxic-comment-classification-challenge/data}} It was fine-tuned for predicting 6 labels related to toxicity: \textit{toxicity}, \textit{severe toxicity}, \textit{obscene}, \textit{threat}, \textit{insult}, and \textit{identitiy attack}, achieving an average Area Under the ROC Curve (\emph{AUC}) score of 98.6\%. For the \textit{toxicity} label only, the classifier achieved 0.82 macro and 0.93 weighted F1 scores (the dataset is imbalanced). This classifier is applied to the original test set comprising 153,164 texts, and predicted the toxicity label for 36,148 texts (23.6\% of the test set).\footnote{The ground truth labels are available only for a subset of the test set (42\%).}

\subsection{Component evaluation: Explainer}
\label{ssec:explainer-evaluation}
The \textit{Explainer} aims to identify the most crucial words utilized by the classifier for predictions (as described in \S \ref{sec:sec-3-our-framework}-A). The employed XAI technique can influence the words recognized as significant, thereby impacting the identified protected attributes on which the model relies to make predictions.

\label{ssec:explainer-evaluation}
\subsubsection{Evaluation metrics}  To evaluate the effectiveness of the \textit{Explainer} component, we first measure the impact on the model's predictive performance (F1 score) by removing the most important words identified by each XAI technique. An effective and precise explainer should result in a noticeable decrease in the predictive performance when these words are removed. Secondly, to assess the sensitivity of the \textit{Explainer}, we measure the overlap of the most important words identified by different XAI techniques. A substantial overlap indicates consistent outputs across XAI techniques. Lastly, we measure the computation time for generating explanations to assess the efficiency of the \textit{Explainer} based on the XAI technique, which is a crucial aspect when dealing with large datasets. 

\subsubsection{Explainer setup} There are two main categories of XAI techniques to compute explanations within a sentence: permutation-based and gradient-based \cite{danilevsky-etal-2020-survey}. For this comparison, we instantiated the \textit{Explainer} component with \textit{SHapley Additive exPlanations} (SHAP) \cite{NIPS2017_7062} as a representative of the permutation-based and with \textit{Integrated Gradients} (IG) \cite{10.5555/3305890.3306024} of the gradient-based techniques. Both techniques have demonstrated competitive performance in prior studies \cite{atanasova-etal-2020-diagnostic}. Specifically, for SHAP, we used the \textit{text permutation explainer} with 3,000 as the maximum evaluation step parameter. For Integrated Gradients, we exploited the implementation provided by the Ferret~\cite{attanasio-etal-2023-ferret} library.

\subsubsection{Results} 
\begin{figure*}
    \centering
   \includegraphics[width=0.55\textwidth]{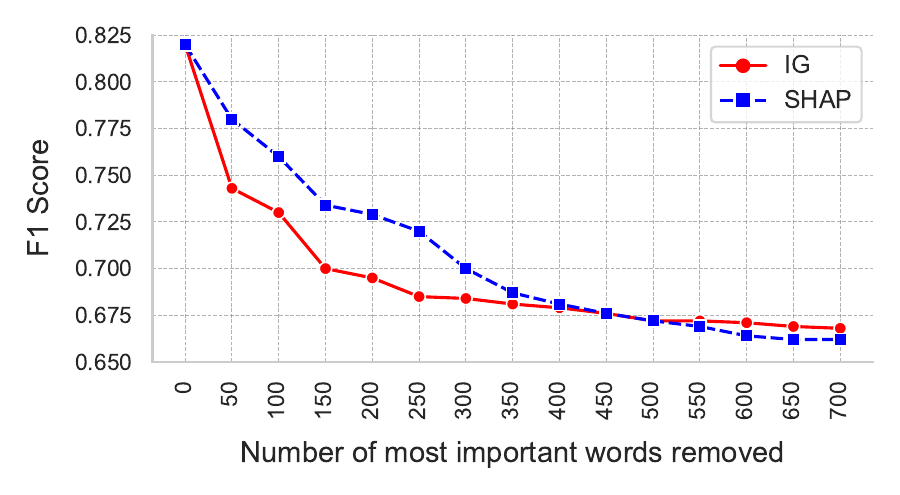}
    \vspace{-5mm}
    \caption{\textbf{Explainer component evaluation.} F1 score decrease by removing the most important words from the test set, extracted by the \textit{Explainer} component with Integrated Gradients (IG) and SHAP techniques. A greater decrease indicates a higher precision in identifying the most important words for predictions.}
    \label{fig:XAI-comparison}
\end{figure*}

We produced the explanations within each sentence over the toxic texts in the test set using both techniques (SHAP and IG); we then aggregated individual scores and extracted an ordered list of the most toxic words as previously described in \S \ref{sec:sec-3-our-framework}\--A. Figure~\ref{fig:XAI-comparison} shows the decrease in the F1 score by removing the most important words in the range from $50$ to $700$ with a step of 50. As expected, removing the most important words from the test set causes a marked decrease in predictive performance, especially for the top 250 words. IG exhibits higher precision, leading to a more substantial decrease initially. However, the decrement tends to converge on the top 400 words for both techniques. This shows that both techniques effectively extract the words used by the classifier for its predictions. We selected the top 400 most toxic words (approximately 10\%) -- removing additional words caused a lower decrease in predictive performance.

We then measured the overlap between the 400 most toxic words identified with IG and SHAP. We found that 307 out of the 400 words were identical (77\%), indicating a substantial agreement between the two techniques. The 23\% of disagreement may be attributable to the varying precision levels inherent in the two methods. 

Finally, we compared the execution times required to generate explanations using both techniques. We observed that IG significantly outperformed SHAP, completing the explanation process more than two orders of magnitude faster. On average, the execution time in seconds to obtain an explanation is approximately 0.2 for IG and 30 for SHAP, using a single Nvidia RTX A6000 GPU.

In summary, Integrated Gradients proves to be the most effective technique for the \textit{Explainer} component, as it exhibits higher precision in identifying the most crucial words and executes significantly faster. As a result, we adopt Integrated Gradients in the \textit{Explainer} for all subsequent experiments. Nevertheless, the framework allows the use of alternative XAI techniques.

\subsection{Component evaluation: Identifier}
\label{ssec:protected-attribute-identifier-evaluation}

The \textit{Identifier} aims to determine which of the most important words are actually protected attributes (\S \ref{sec:sec-3-our-framework}-B). To evaluate its effectiveness, we compare the protected attributes identified by the component instantiated with human-in-the-loop and machine-in-the-loop approaches against the annotations provided by two expert annotators, who possess a greater depth of knowledge of the definitions of protected categories by AI regulators than participants engaged in the human study.\footnote{Expert annotators are people within our team with a background in human-computer interaction and trustworthy and responsible AI. They are located in two different Western countries, with different ethnicities and ages. They carefully read the UK Equality Act 2010 and unanimously agreed on the annotation to be performed, which was done independently.} We also add for comparison a pre-defined dictionary of 51 protected attributes from previous works \cite{10.1145/3308558.3313504,10.1145/3278721.3278729}. 

\subsubsection{Evaluation metrics} We measure the Cohen’s kappa inter-annotator agreement \cite{doi:10.1177/001316446002000104} to evaluate the accuracy and reliability of the protected attributes identified by the different approaches. It ranges between $[0,1]$; the higher the score is, the higher the agreement.

\subsubsection{Protected attributes identifier setup} We selected the 400 most toxic words extracted with the \textit{Explainer} component instantiated with Integrated Gradients as the candidate set to identify protected attributes. Then, we configured the \textit{Identifier} as follows.

\vspace{1mm}
\noindent 
\textbf{Human-in-the-loop setup.} We set up an MTurk study where we asked annotators to label words with the protected category they refer to (if any). As anticipated in \S \ref{sec:sec-3-our-framework}-B1, we also used a trap mechanism to detect poor quality responses; specifically, we gave annotators the following definition of toxicity: \textit{``Toxic language is a way of communicating that harms other people''}. Then, for each word, we also asked participants to answer the following additional trap question:
\begin{description}
    \item \footnotesize \textbf{Trap Question:} \textit{Does the word \textcolor{airforceblue}{\textsc{\{word\}}} suggest toxic language?} 
    \item \footnotesize \textbf{Possible Answers:} 1. \textit{Not at all}, 2. \textit{Very little}, 3. \textit{Somewhat}, 4. \textit{To a great extent}, 5. \textit{Definitely}.
\end{description}
To the list of 400 most toxic words, we added 15 trap words that can be easily classified as \textit{toxic} (e.g., ``asshole'') or \textit{non-toxic} (e.g., ``friendly''). The full list of trap words can be found in Table~\ref{tab:trap-words} in Appendix B. For the \textit{non-toxic} (\textit{toxic}) trap words, we expected MTurk participants to select a score of 1 or 2 (4 or 5) on the Likert scale. Participants were considered unreliable if they did not meet those expectations, and their assessments were discarded from our results. We ended up with 246 reliable participants, evenly split between males and females. The majority were educated (74\% finished college), mostly located in the United States, falling within the median age group of 26-39. In terms of racial demographics, most were White (52\%), followed by Asian (27\%), African (7\%), and Hispanic (5\%).\footnote{Crowdworkers have been paid for their valuable contributions and time devoted to this research.}

We collected five annotations per word on average. A word was labeled as a protected attribute if the sum of the votes across the nine categories exceeded the \textit{None of the above} (majority voting).

\vspace{1mm}
\noindent 
\textbf{Machine-in-the-loop setup.} We also annotated the same set of words by prompting GPT-3.5-Turbo, as introduced in \S \ref{sec:sec-3-our-framework}-B2. The temperature parameter was set to 0.3 to limit creativity in generating the responses. Other temperature values in the range $[0.3, 0.7]$ have been experimented with, although no major differences were observed. For each candidate word, we prompted GPT-3.5-Turbo, asking for a possible protected category, a reliability score, and an explanation for the classification. If a word is classified with any protected category, it is labeled as a protected attribute.

\begin{figure*}
    \centering
   \includegraphics[width=0.45\textwidth]{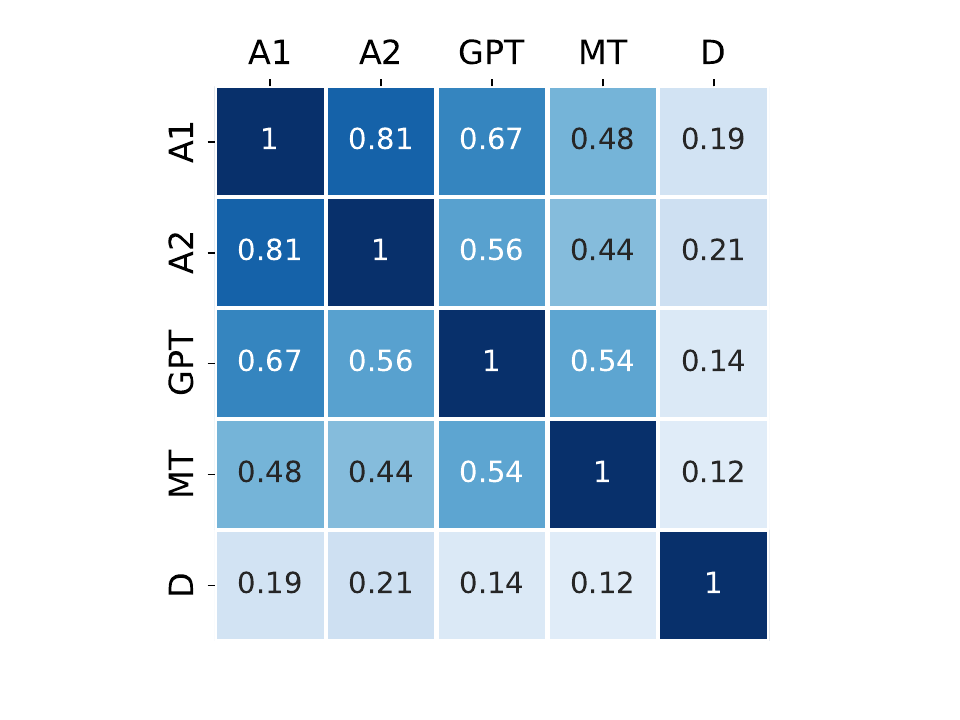}
    \vspace{-5mm}
    \caption{\textbf{Identifier component evaluation.} Cohen’s kappa annotator agreement in labeling protected attributes for the 400 most toxic words. The annotation was performed by two expert annotators (A1 and A2), ChatGPT (GPT), MTurk (MT), and a pre-defined dictionary (D). The two-by-two Cohen’s kappa annotator agreement is reported, in the range [0, 1], where a higher score indicates a higher level of agreement.}
    \label{fig:annotator-agreement}
\end{figure*}

\subsubsection{Results}  The two expert annotators (A1, A2) identified 72/400 (18\%) and  66/400 (17\%) protected attributes, respectively. The human-in-the-loop (MTurk) approach 108/400 (27\%) ones. Instead, the machine-in-the-loop (ChatGPT) approach labeled 93/400 (23\%) words as protected attributes. These findings indicate that the original classifier heavily relies on protected attributes for toxicity predictions.
Interestingly, the ChatGPT \textit{Identifier} is also able to annotate proxy words for the protected categories. For example, the word \textit{`headscarf'} is annotated as related to \textit{religion and belief} (see Figure~\ref{fig:prompt5} in Appendix A). If we use the pre-defined dictionary \cite{10.1145/3308558.3313504,10.1145/3278721.3278729} instead of the \textit{Identifier} component, only 9 out of 400 words (2\%) would have been labeled as protected attributes. This suggests that pre-defined dictionaries may consist of a limited subset of protected attributes and not encompass the entire range of relevant attributes.

Figure~\ref{fig:annotator-agreement} shows the two-by-two Cohen’s kappa inter-annotator agreement between the annotation performed by the two experts (A1, A2), ChatGPT (GPT), MTurk (MT), and the pre-defined dictionary (D). A higher score indicates a greater level of agreement. The score between the two expert annotators is 0.81, corresponding to an almost perfect agreement according to Landis and Koch’s scale \cite{landis1977measurement}. The ChatGPT annotations demonstrate substantial (0.67) and moderate (0.56) agreement with the expert annotators. In contrast, the MTurk annotations show a moderate agreement of 0.48 and 0.44 with the expert annotators, respectively. The pre-defined dictionary exhibits low agreement with all other annotations, as it only covers a small subset of the important words.

This evaluation demonstrates that the machine-in-the-loop approach outperforms the human-in-the-loop one in identifying protected attributes, while also enabling the full automation of the framework. For the next experiments, we thus adopt the LLM-based approach for the \textit{Identifier}.

\subsection{Component evaluation: Moderator}
\label{ssec:moderator-component-evaluation}
The \textit{Moderator} aims to create a mitigated training corpus to train a new classifier with reduced reliance on protected attributes and similar predictive performance. To evaluate the effectiveness of each mitigation strategy, we trained and evaluated a distinct mitigated model for each strategy.

\label{ssec:moderator-evaluation}
\subsubsection{Evaluation metrics}  For each mitigation strategy, we examine two key aspects of the mitigated classifiers: \textit{fairness} and \textit{predictive performance}. 
Our \textit{fairness} is defined as {\em fairness through unawareness}, whereby ``an algorithm is fair as long as any protected attributes are not explicitly used in the decision-making process'' \cite{10.1145/3457607}. This is quantified by measuring the number of protected attributes each mitigated model relies on in making predictions based on the \textit{Explainer} and \textit{Identifier} components.
A lower number indicates a reduced dependence on protected attributes, which signifies progress towards a more fair and unbiased classifier. To evaluate the \textit{predictive performance}, we measure the F1 score specifically for the toxicity label, providing insight into the model's accuracy in identifying toxic texts.
Additionally, we evaluate the Area Under the Curve (\emph{AUC}) score for all toxicity-related labels. This metric provides an overall measure of the model's performance in identifying various aspects of toxicity. By considering both \textit{fairness} and \textit{predictive performance}, we can ascertain the effectiveness of the mitigated models in achieving a balance between reducing reliance on protected attributes and maintaining similar predictive capabilities.

\subsubsection{Moderator setup} For the mitigation strategies outlined in \S\ref{sec:sec-3-our-framework}-C, we used the following setup: for the removal-based strategies (MS1, MS2), we removed the sentences or the words if, after the tokenization, the protected attributes are in the list of tokens. In the case of the mitigation strategies based on the $k$-neighbours (MS3, MS4), we set the value of $k$ to 5, meaning that for each protected attribute, the five closest words were identified. To identify these nearest neighbors, we computed the cosine similarity between each word in the vocabulary and the protected attribute using the 300-dimensional GloVe \cite{pennington2014glove} word embedding, as suggested in \cite{10.1145/3308558.3313504}. For the hypernyms-based strategy (MS5), we utilized the WordNet lexical database \cite{10.1145/219717.219748} provided in NLTK,\footnote{\url{https://www.nltk.org/howto/wordnet.html}} as suggested in~\cite{10.1145/3308558.3313504}. We replaced each protected attribute with its first-level hypernym extracted from its synset of synonyms.

\subsubsection{Training the mitigated models}
We applied each mitigation strategy to the original Wikipedia comments training dataset. Each mitigation strategy produced a modified version of the training dataset (containing 159,571 examples), whose differences are shown in the third column of Table~\ref{tab:results-in-distribution}. The sentence-removal mitigation strategy (MS1) resulted in a decrease of 6k examples. Instead, the strategy that added $k$ new sentences for each protected attribute (MS4) increased the training dataset by 108k new sentences. The other mitigation techniques did not change the number of training examples. All mitigated models ($M_{1}^{*}$-$M_{5}^{*}$) were trained by fine-tuning the original pre-trained weights of BERT\footnote{\url{https://huggingface.co/bert-base-uncased}} for 3 epochs, with a batch size of 16, and Adam \cite{kingma2014adam} as optimizer. 

To evaluate the mitigated models, we first classified all texts in the test set with each \textit{mitigated} model. Then, we applied the \textit{Explainer} to extract the most important predictive words used by each \textit{mitigated} model for the toxicity predictions in those texts. Finally, we exploited the \textit{Identifier} to determine if the new important words of the mitigated models were protected attributes.

\subsubsection{Results} 

\begin{table}[]
\caption{\textbf{Results in mitigating toxicity prediction on in-distribution data (test set).}
The original model is highlighted in grey ($M_o$). For each mitigated model are reported: (1) the model identifier, (2) the mitigation strategy applied, (3) the difference in training examples after the mitigation strategy, (4) the F1 macro and weighted scores for the toxicity class on the original test set, (5) the \emph{AUC} score for all toxicity-related labels on the original test set, and (6) the percentage and ratio of relied-upon protected attributes, with the number of those present in the original model in curly brackets. 
The best performing for each metric is in bold.}
\label{tab:results-in-distribution}
\vspace{-0.1in}
\scriptsize
\begin{tabular}{cllc|ccccc|ccc}
\toprule
   &  &                                                &                                                                                          & \multicolumn{5}{c|}{\textit{Predictive Performance} $\uparrow$}                                                                                                                                                        & \multicolumn{3}{c}{\textit{Fairness} $\downarrow$}                              \\
\multirow{-2}{*}{\textbf{\begin{tabular}[c]{@{}c@{}}Model \\ ID\end{tabular}}} &  & \multirow{-2}{*}{\textbf{Mitigation Strategy}} & \multirow{-2}{*}{\textbf{\begin{tabular}[c]{@{}c@{}}$\Delta$ Train\\ Examples\end{tabular}}} & \textbf{\begin{tabular}[c]{@{}c@{}}F1 \\ Macro\end{tabular}} & \multicolumn{1}{l}{\textbf{}} & \textbf{\begin{tabular}[c]{@{}c@{}}F1 \\ Weight\end{tabular}} & \multicolumn{1}{l}{\textbf{}} & \textbf{AUC} & \textbf{\% PA} & \multicolumn{1}{l}{\textbf{}} & \textbf{Ratio PA} \\ \midrule
\rowcolor[HTML]{EFEFEF} 
$M_{o}$  &  & \multicolumn{1}{c}{\cellcolor[HTML]{EFEFEF}-}  & -                                                                                        & 0.815                                                        &                               & 0.926                                                         &                               & \textbf{0.986}        & 23\%           &                               & 93/400            \\  \midrule
$M_{1}^{*}$    &  & \textit{MS1 - Sentence removal}                & -6k                                                                                      & 0.816                                                        &                               & 0.934                                                    &                               & 0.981        & \textbf{9\%}            &                               & \textbf{37/400 \{16\}}   \\
$M_{2}^{*}$                                                                             &  & \textit{MS2 - Word removal}                    & -                                                                                        & \textbf{0.828}                                                        &                               & \textbf{0.938}                                                         &                               & 0.981        & 10\%           &                               & 40/400 \{21\}   \\
$M_{3}^{*}$                                                                            &  & \textit{MS3 - Word replace 1 rand syn}     & -                                                                                        & 0.812                                                        &                               & 0.926                                                         &                               & 0.983        & 14\%           &                               & 56/400 \{36\}   \\
$M_{4}^{*}$                                                                             &  & \textit{MS4 - Word replace k rand syn}             & +108k                                                                                    & 0.783                                                        &                               & 0.908                                                         &                               & 0.981        & 20\%           &                               & 80/400 \{50\}   \\
$M_{5}^{*}$                                                                             &  & \textit{MS5 - Word replace hyper}           & -                                                                                        & 0.812                                                        &                               & 0.927                                                         &                               & 0.983        & 13\%           &                               & 51/400 \{32\}   \\ \bottomrule
\end{tabular}
\end{table}

\vspace{1mm}
\noindent
\\ \textbf{Fairness.} The last two columns in Table~\ref{tab:results-in-distribution} show the percentage and number of the most toxic words labeled as protected attributes for all the mitigated models ($M_{1}^{*}$-$M_{5}^{*}$). The number of those already present among the protected attributes of the original model ($M_o$) is indicated in curly brackets. All the mitigation strategies reduced the number of protected attributes the model relied upon. However, the mitigated models trained with removal-based strategies (MS1, MS2) achieved much better results. Only 9\% and 10\% of their most toxic words were labeled as protected attributes (37 and 40 words out of 400), representing a decrease of 61\% from the original model (93 words out of 400). One possible reason for the lower performance of replacement-based mitigation strategies (MS3-MS5) is that they can introduce new protected attributes when replacing words. For a qualitative evaluation of fairness improvement, please refer to Figure~\ref{tab:motivation-example-mitigated-model} and Figure~\ref{fig:motivation-example-mitigated-model} in Appendix C.

\vspace{1mm}
\noindent
\textbf{Predictive performance.} In Table~\ref{tab:results-in-distribution}, columns 4 and 5 report the macro and weighted F1 scores on the toxicity label for the original and mitigated models. Column 6 also presents the mean \emph{AUC} scores across all the toxicity-related labels. The results show that all the mitigated models present similar F1 scores compared to the original model, except for the one trained with MS4, which exhibits a greater decrease. Interestingly, the mitigated models trained on the removal-based mitigation strategies (MS1, MS2) achieve better F1 scores than the original model. The word-removal (MS2) increases the macro and weighted F1 scores by 1.3\% and 1.2\%. Indeed, we observed that the removal-based mitigation strategies reduced the number of false positives in the toxicity predictions (i.e., non-toxic texts wrongly predicted as toxic). All the mitigated models exhibit slightly lower \emph{AUC} scores compared to the original model. However, the decrease is minor and acceptable (around 0.5\% and 0.3\%) in light of the reduced reliance on protected attributes.

\vspace{1mm}
\noindent
\textbf{Summary.} We conclude that our framework effectively reduces the model's reliance on protected attributes without compromising its predictive performance. Indeed, all mitigated models are fairer in that they significantly reduce the use of protected attributes and exhibit similar predictive performance to the original model. Interestingly, the removal-based strategies (MS1, MS2) even increased the models' predictive performance after the mitigation.

\section{Framework Evaluation: Generalizability}
\label{sec:results-general}
We finally evaluate the generalizability of our framework first to toxicity prediction on out-of-distribution data (\S \ref{ssec:evaluation-outdistribution}), and second on different tasks, i.e., sentiment analysis (\S \ref{ssec:evaluation-sentiment}) and occupation classification (\S \ref{ssec:evaluation-occupations}). 

\subsection{Framework and evaluation settings} For this evaluation, we instantiated the \textit{Explainer} with Integrated Gradients, the \textit{Identifier} with ChatGPT, and the \textit{Moderator} with the removal-based mitigation strategies. As shown in \S \ref{sec:results-performance}, this turned out to be the optimal framework configuration. We perform a similar evaluation of the mitigated models by measuring their \textit{fairness} and \textit{predictive performance}. \textit{Fairness} is evaluated by quantifying the number of protected attributes each mitigated model relies on ({\em fairness through unawareness}).
\textit{Predictive performance} is evaluated using quantitative metrics on the test set.
For the toxicity classifier, we measure the F1 score for the toxicity label, which allows us to gauge the model's accuracy in detecting instances of toxicity, and the Area Under the Curve (\emph{AUC}) score for all toxicity-related labels, providing an overall assessment of the model's performance in identifying different aspects of toxicity. For the sentiment and the occupation classifiers, we solely measure the F1 score, as it provides a comprehensive assessment of the model's accuracy in these tasks.

\subsection{Mitigating toxicity prediction on out-of-distribution data}
\label{ssec:evaluation-outdistribution}
This experiment aims to assess the applicability of our framework in mitigating the toxicity model when applied to out-of-distribution data, specifically company reviews. This is crucial as classifiers are normally applied to datasets from other domains with different word distributions from training.

\subsubsection{Company reviews data} We collected data from a popular online platform where current and former employees write reviews about companies. Reviewers comment on various aspects such as personal experience with the company or managers, salary information, workplace culture, and typical job interviews.  The platform fosters a constructive approach among its users by manually and automatically moderating the content of reviews. However, reviews are published anonymously. On the one hand, this promotes user privacy. On the other hand, it can also cause some users to write public insults and offenses toward companies or people. Specifically, we collected a dataset of 439,163 reviews from U.S.-based companies across all 51 U.S. states written from 2008 to 2020.\footnote{To preserve the privacy of individuals, Personally Identifiable Information (PII) was removed.} 
Each review contains a \textit{pros} part (positive comments in the review) and a \textit{cons} part (negative comments). We applied the same toxicity classifier introduced in \S \ref{ssec:in-distribution-task} to identify toxic company reviews.

\subsubsection{Toxicity in company reviews} The initial expectation was not to have many toxic reviews in the dataset due to the highly curated nature of the platform. However, if we consider a post to be \textit{toxic} when at least one of the \textit{cons} or \textit{pros} fields contains inappropriate content, we found $1.6\%$ of reviews (7,224) to be toxic. The number of reviews classified as toxic by using the \textit{pros} and \textit{cons} texts as input is 853 for \textit{pros} ($0.2\%$) and 6,495 for \textit{cons} ($1.5\%$) over 439,163. As expected, we found that most of the toxic texts are present in \textit{cons}. Interestingly, some people tend to be so angry and frustrated by the work experience that they let off steam even in the \textit{pros} field. 

\subsubsection{Identify protected attributes in toxicity predictions on company reviews} All \textit{pros} and \textit{cons} reviews predicted as \textit{toxic} were analyzed by the \textit{Explainer} component to extract the most important words used by the model in predicting toxic reviews. Then, we selected the 400 most toxic words extracted, and we annotated those words with GPT-3.5-Turbo. Among the 400 most important words used by the model in predicting toxic reviews, 76 are protected attributes (19\%), as shown in the last two columns of the first row in Table~\ref{tab:results-out-distribution} (original model $M_o$). We can conclude that the original classifier exhibits a significant reliance on protected attributes for toxicity predictions, even when applied to different out-of-distribution data. 

\subsubsection{Training the mitigated models} We applied the removal-based mitigation strategies (MS1, MS2) to the original Wikipedia comments training dataset based on the protected attributes identified in the toxic company reviews. Table~\ref{tab:results-out-distribution} shows the differences in the number of training examples after each strategy in the third column. The original training dataset contained 159,571 examples. MS1 resulted in a decrease of 6k examples, while MS2 did not change it. All mitigated models were fine-tuned for 3 epochs, with a batch size of 16, and Adam as optimizer. To evaluate the mitigated models, all \textit{pros} and \textit{cons} reviews were classified by each \textit{mitigated} model. Then, we applied the \textit{Explainer} component to extract the most important 400 predictive words used by each \textit{mitigated} model for the toxicity predictions on company reviews. Finally, we exploited the \textit{Identifier} to determine if the new important words of the mitigated models were protected attributes.

\begin{table}[]
\caption{\textbf{Results in mitigating toxicity prediction on out-of-distribution data (company reviews).}
The original model is highlighted in grey ($M_o$). For each mitigated model are reported: (1) the model identifier, (2) the mitigation strategy applied, (3) the difference in training examples after the mitigation strategy, (4) the F1 macro and weighted scores for the toxicity class on the original test set, (5) the \emph{AUC} for all toxicity-related labels on the original test set, (6) the percentage and ratio of relied-upon protected attributes, with the number of those present in the original model in curly brackets. 
The best performing for each metric is in bold. }
\label{tab:results-out-distribution}
\vspace{-0.1in}
\scriptsize
\begin{tabular}{cllc|ccccc|ccc}
\toprule
&  &                                                &                                                                                          & \multicolumn{5}{c|}{\textit{Predictive Performance} $\uparrow$}                                                                                                                                                          & \multicolumn{3}{c}{\textit{Fairness} $\downarrow$}                                    \\
\multirow{-2}{*}{\textbf{\begin{tabular}[c]{@{}c@{}}Model \\ ID\end{tabular}}} &  & \multirow{-2}{*}{\textbf{Mitigation Strategy}} & \multirow{-2}{*}{\textbf{\begin{tabular}[c]{@{}c@{}}$\Delta$ Train\\ Examples\end{tabular}}} & \textbf{\begin{tabular}[c]{@{}c@{}}F1 \\ Macro\end{tabular}} & \multicolumn{1}{l}{\textbf{}} & \textbf{\begin{tabular}[c]{@{}c@{}}F1 \\ Weight\end{tabular}} & \multicolumn{1}{l}{\textbf{}} & \textbf{AUC}   & \textbf{\% PA} & \multicolumn{1}{l}{\textbf{}} & \textbf{Ratio PA}       \\ \midrule
\rowcolor[HTML]{EFEFEF} 
$M_{o}$   &  & \multicolumn{1}{c}{\cellcolor[HTML]{EFEFEF}-}  & -                                                                                        & 0.815              &                               & 0.926                                                         &                               & \textbf{0.986} & 19\%           &                               & 76/400                  \\ \midrule
$M_{1}^{*}$              &  & \textit{MS1 - Sentence removal}                & -6k                                                                                      & 0.824                                                        &                               & \textbf{0.938}                                                &                               & 0.979          & \textbf{4\%}   &             & \textbf{16/400 \{8\}} \\
$M_{2}^{*}$                   &  & \textit{MS2 - Word removal}                    & -                                                                                        & \textbf{0.825}                         &                               & 0.935                           &                               & 0.983          & 5\%            &                               & 19/400 \{11\}         \\ \bottomrule
\end{tabular}
\end{table}

\subsubsection{Results} 
\vspace{1mm}
\noindent
\\ \textbf{Fairness.} The last two columns in Table~\ref{tab:results-out-distribution} show the percentage and number of the most toxic words labeled as protected attributes. The number of those already present among the protected attributes used by the original model ($M_o$) is also indicated in curly brackets. The results confirm that removal-based mitigation strategies reduce the number of protected attributes the model relied upon. MS1 and MS2 reduce the percentage of protected attributes from 19\% to 4\% and 5\% (16 and 19 out of 400 words), respectively. This corresponds to a decrease of 79\% and 75\%. They also reduce the protected attributes the original classifier relies on from 76 to 8 and 11, respectively.

\vspace{1mm}
\noindent
\textbf{Predictive performance.} Columns 4 and 5 in Table~\ref{tab:results-out-distribution} show the macro and weighted F1 scores achieved by the original and mitigated models on the test set. The mitigated models exhibit higher predictive performance in terms of F1 scores than the original model. The increment is around 1\% for both scores and mitigated models. Finally, column 6 shows the \emph{AUC} for all the toxicity-related labels. The mitigated model produced by MS1 achieves 0.979 on the \emph{AUC} score, with a decrease of 0.007 from the original model. Instead, with MS2, the decrease in performance is only 0.003.

\vspace{1mm}
\noindent
\textbf{Summary.} The experimental results obtained from the out-of-distribution data demonstrate the capability of our framework to effectively mitigate a model's reliance on protected attributes when applied to non-training data, where ground truth labels are unavailable. It showcases its adaptability and robustness in real-world scenarios where labeled data may not be readily accessible.

\subsection{Mitigating sentiment analysis}
\label{ssec:evaluation-sentiment}
This evaluation aims to assess the versatility and effectiveness of our framework across different classification tasks. For this experiment, we chose sentiment classification to test our framework in mitigating the use of protected attributes for tasks where their reliance might be lower.

\subsubsection{Training the original sentiment classifier}
We selected a dataset of 163K tweets and 37K Reddit comments in English, expressing people's opinions towards the general elections held in India in 2019.\footnote{\url{https://www.kaggle.com/datasets/cosmos98/twitter-and-reddit-sentimental-analysis-dataset}} The task consists of a multi-class sentiment classification problem with 3 classes: \textit{negative}, \textit{neutral}, and \textit{positive}. We split the dataset with 80\% for training (160k) and 20\% for testing (40k). We fine-tuned the BERT model for 3 epochs, achieving a 0.96 F1 score on the test set.

\subsubsection{Identifying protected attributes in sentiment predictions}
We used the fine-tuned model to predict the sentiment label over the entire test set. Then, we analyzed, separately, all the \textit{negative} and \textit{positive} texts with the \textit{Explainer} component instantiated with Integrated Gradients. The \textit{neutral} texts do not contain specific patterns that the model should learn and are not of interest for mitigation. Then, we annotated with GPT-3.5-Turbo the 5\% of the most important words for the \textit{negative} and 5\% for the \textit{positive} texts separately, resulting in the top 200 negative and 200 positive words. We found that 16 (8\%) of 200 negative words and 11 (6\%) of 200 positive words were labeled as protected attributes by the Identifier, suggesting a moderate reliance on protected attributes.

\subsubsection{Training the mitigated models} We applied the two removal-based mitigation strategies (MS1, MS2). In this case, the mitigation is performed separately per class label (e.g., the protected attributes in the most negative words are mitigated only on the negative training examples). MS1 decreases the training set by 5k \textit{negative} and 8k \textit{positive} training examples, as shown in the third and fourth columns in Table~\ref{tab:results-sentiment-analysis}. Also in this case, the models were fine-tuned for 3 epochs.

We used the \textit{mitigated} models to predict the sentiment label over the test set. Then, we extracted the most important 200 words from the \textit{negative} and \textit{positive} texts separately with the \textit{Explainer}, and we annotated those words with the \textit{Identifier}.

\subsubsection{Results} 

\begin{table}[]
\caption{\textbf{Results in mitigating sentiment analysis.} The original model is highlighted in grey ($M_o$). For each mitigated model are reported: (1) the model identifier, (2) the mitigation strategy applied, (3) the difference in training examples after the mitigation strategy for the \textit{negative} and \textit{positive} classes, (4) the F1 macro score on the original test set, and (5) the percentage and ratio of relied-upon protected attributes, with the number of those present in the original model in curly brackets. 
The best performing for each metric is in bold.}
\label{tab:results-sentiment-analysis}
\vspace{-0.1in}
\scriptsize
\begin{tabular}{cllcc|c|cclcc}
\toprule
   &  &                                                & \multicolumn{2}{c|}{}                                                                                          &                                                                                              & \multicolumn{5}{c}{\textit{Fairness} $\downarrow$}                                                        \\
   &  &                                                & \multicolumn{2}{c|}{\multirow{-2}{*}{\textbf{\begin{tabular}[c]{@{}c@{}}$\Delta$ Train \\ Examples\end{tabular}}}} & \multirow{-2}{*}{\textit{\begin{tabular}[c]{@{}c@{}}Predictive \\ Performance\end{tabular}} $\uparrow$} & \multicolumn{2}{c}{\textit{Negative Class}} &  & \multicolumn{2}{c}{\textit{Positive Class}} \\
\multirow{-3}{*}{\textbf{\begin{tabular}[c]{@{}c@{}}Model \\ ID\end{tabular}}} &  & \multirow{-3}{*}{\textbf{Mitigation Strategy}} & \multicolumn{1}{l}{\textit{Negative}}                            & \textit{Positive}                           & \textbf{F1}                                                                                  & \textbf{\% PA}   & \textbf{Ratio PA}        &  & \textbf{\% PA}   & \textbf{Ratio PA}        \\ \midrule
\rowcolor[HTML]{EFEFEF} 
$M_{o}$                                                                            &  & \multicolumn{1}{c}{\cellcolor[HTML]{EFEFEF}-}  & -                                                                & -                                           & 0.96                                                                                         & 8\%              & 16/200                   &  & 6\%              & 11/400                   \\ \midrule
$M_{1}^{*}$                                                                             &  & \textit{MS1 - Sentence removal}                & -5k                                                              & -8k                                         & 0.96                                                                                         & \textbf{4\%}     & \textbf{8/200 \{2\}}   &  & \textbf{3\%}     & \textbf{5/200 \{1\}}   \\
$M_{2}^{*}$                                                                            &  & \textit{MS2 - Word removal}                    & -                                                                & -                                           & 0.96                                                                               & 8\%              & 16/200 \{2\}           &  & 6\%              & 11/200 \{2\}           \\ \bottomrule
\end{tabular}
\end{table}

\vspace{1mm}
\noindent
\textbf{\\ Fairness.} The last four columns in Table~\ref{tab:results-sentiment-analysis} show the percentage and the number of protected attributes of the original ($M_o$) and mitigated ($M_{1}^{*}$ and $M_{2}^{*}$) sentiment classifiers for the \textit{negative} and \textit{positive} classes, separately. MS1 produces a mitigated model that relies on half of the protected attributes of the original classifier (4\% for the \textit{negative} and 3\% for the \textit{positive} classes). Interestingly, the number of protected attributes the original classifier relied on is almost completely mitigated, except for 2 for the \textit{negative} and 1 for the \textit{positive} classes. MS2 has a similar behavior in this. However, many new protected attributes emerge as new important words. In the end, the total number of protected attributes remains the same, even though the protected attributes of the original model have almost all been mitigated. Therefore, MS1 is the most effective in this case.

\vspace{1mm}
\noindent
\textbf{Predictive performance.} The fifth column in Table~\ref{tab:results-sentiment-analysis} shows the F1 score obtained by the original sentiment classifier and the mitigated models on the test set. The mitigated models achieve the same F1 score, thus showing the same predictive capabilities.

\vspace{1mm}
\noindent
\textbf{Summary.} These results show that our framework can be effective not only on the toxicity model, which heavily relies on protected attributes, but also on the sentiment model, which is moderately impacted by protected attributes, confirming its general effectiveness.

\subsection{Mitigating occupation classification}
\label{ssec:evaluation-occupations}
This evaluation serves three primary objectives: (1) to further assess the adaptability and effectiveness of our framework across various classification tasks, (2) to mitigate the use of protected attributes in scenarios where the final prediction has tangible consequences for individuals, and (3) to compare our framework with two mitigation techniques that act on the model rather than data.

To achieve these goals, we selected the task of predicting occupations from online biographies, using a dataset of biographies annotated by gender and occupation from previous works~\cite{10.1145/3287560.3287572,ravfogel-etal-2020-null,Peng_Nushi_Kıcıman_Inkpen_Suri_Kamar_2019}. We used the field \textit{`cleaned\_input\_text'} as input text, where sentences that directly reveal the occupation were removed (e.g., ``he is a journalist''), and we removed first names. We compare our framework with two model-based mitigation techniques: \textit{Iterative Null-space Projection (INLP)} \cite{ravfogel-etal-2020-null} and \textit{Entropy-based Attention Regulation (EAR)} \cite{attanasio-etal-2022-entropy}. \textit{INLP} requires an additional annotation for the mitigated category, which is only present for gender in this dataset. Therefore, we conduct two distinct analyses: (1) focusing solely on gender-related, and  (2) examining all protected categories together. For the gender-related protected attributes, we compare both baselines with our word-removal (MS2). Instead, we use only \textit{EAR} as a baseline to mitigate all the protected categories. We chose MS2 because it has shown similar mitigation effectiveness while maintaining competitive performance, but it has higher flexibility across datasets than sentence-removal (MS1) (see \S\ref{ssec:limitations}).

\subsubsection{Training the original occupation classifiers}
The dataset contains 393,423 biographies for 28 occupations split into 255,710, 39,369, and 98,344 train, dev, and test examples. We fine-tuned a BERT-base and uncased model for each occupation in one-vs-all settings (i.e., a binary model that predicts the occupation or not for each label). Due to the high imbalance of the dataset, we performed the experiments for the five most frequent classes (i.e., \textit{nurse}, \textit{attorney}, \textit{journalist}, \textit{physician}, and \textit{professor}). The models were fine-tuned for 3 epochs.\footnote{We utilized inversely proportional class weights in the loss function due to the highly imbalanced training dataset.} 
The second column in Table~\ref{tab:results-occupacies} shows the macro F1 score on the test set for each original model trained for four occupations.\footnote{Results for the \textit{professor} occupation are not reported since the model do not rely on gender-related protected attributes.} Those models are highly effective in classifying occupations, achieving a macro F1 score higher than 0.89. Still, such high performance could be achieved by heavily using protected attributes.

\subsubsection{Identifying protected attributes in occupation classification}
We used each original model to predict each occupation label over the entire test set and analyzed those texts using the \textit{Explainer} to extract the most important words in predicting each occupation. Then, for each occupation, we annotated with GPT-3.5-Turbo the top 400 words to identify protected attributes. We measured the models' reliance on protected attributes related to (1) gender only, and (2) all categories (third and fourth columns in Table~\ref{tab:results-occupacies}). All the models moderately rely on protected attributes. The \textit{nurse} occupation is the most influenced, also by gender-related words, such as pronouns (e.g., \textit{`she'}, \textit{`her'}).

\subsubsection{Training the mitigated models.}
With our framework, we applied the word removal mitigation strategy (MS2) for each occupation on (1) gender-related 
only, and (2) all the protected categories simultaneously. Therefore, we trained two different mitigated models for each occupation. We trained one mitigated model for each occupation with the \textit{INLP} methodology \cite{ravfogel-etal-2020-null}. \textit{INLP} can mitigate the gender-related protected attributes but is not applicable to all the other categories, since the dataset contains the additional annotation only for gender. Specifically, we used the original pre-trained BERT weights as the encoder. Then, we multiplied the embedding representation of the [CLS] token from the last hidden layer for each input text by the projection matrix produced by the \textit{INLP} technique (to ensure that the embedding representation does not encode information about gender). Finally, we added a classification layer on top. We fine-tuned only the classification layer while freezing the BERT encoder and the projection matrix, as suggested in~\cite{ravfogel-etal-2020-null}. For \textit{EAR} \cite{attanasio-etal-2022-entropy}, we used the same BERT architecture, and we added the loss function regularization term, with 0.001 as regularization strength. EAR does not allow the selection of which protected categories to mitigate, but it identifies by itself which words have a high attention entropy. Thus, we trained a single mitigated model, and we evaluated its reliance on gender and all protected categories separately.

We used the mitigated models to predict the occupation labels over the entire test set, we extracted the most important 400 words for each occupation separately with the \textit{Explainer}, and we annotated those words with the \textit{Identifier} to evaluate if they exhibit a reduced reliance on protected attributes.

\begin{table}[]
\caption{\textbf{Results in mitigating occupation classification.} For the \textit{original} models (highlighted in grey), trained in one-vs-all settings, are reported the macro F1 and the reliance on protected attributes (PA) for gender only and all categories. For each occupation, we applied our framework with the word-removal strategy (MS2) on gender-related only and all categories of protected attributes training two different \textit{mitigated} models. For \textit{EAR} \cite{attanasio-etal-2022-entropy}, a single model was trained but evaluated on different protected categories. \textit{INLP} \cite{ravfogel-etal-2020-null} is only applicable to gender-related protected attributes in this dataset. For each mitigated model are reported: (1) the mitigation technique, (2) the macro F1 score for the occupation classification on the test set (\textit{predictive performance}), and (3) the ratio and percentage of relied-upon protected attributes, with the number of those present in the original model in curly brackets (\textit{fairness}). The best performing for each metric is in bold.}
\label{tab:results-occupacies}
\vspace{-0.1in}
\scriptsize
\begin{tabular}{c|ccc|lcccc}
\toprule
\multirow{3}{*}{\textbf{Occupation}}               & \multicolumn{3}{c|}{\textit{Original Model}}                                                         & \multicolumn{1}{c}{\multirow{3}{*}{\textbf{\begin{tabular}[c]{@{}c@{}}Mitigation \\ Technique\end{tabular}}}} & \multicolumn{4}{c}{\textit{Mitigated Models}}                                                                 \\
 & \multirow{3}{*}{\textbf{F1 $\uparrow$}} & \textit{ Gender only} & \textit{All Categories}        & \multicolumn{1}{c}{}                                                                                    &      & \multicolumn{1}{c|}{\textit{Gender only}}          &            & \multicolumn{1}{c}{\textit{All Categories}}  \\
                                              &                                    & \textbf{Ratio (\%) PA $\downarrow$}       & \textbf{Ratio (\%) PA $\downarrow$}       & \multicolumn{1}{c}{}                                                                                          & \textbf{F1 $\uparrow$} & \multicolumn{1}{c|}{\textbf{Ratio (\%) PA  $\downarrow$}}     & \textbf{F1 $\uparrow$} & \textbf{Ratio (\%) PA $\downarrow$} \\ \toprule
\multicolumn{1}{c|}{\multirow{3}{*}{Nurse}}    & \multirow{3}{*}{0.939}             & \multirow{3}{*}{11/400 (3\%)}  & \multirow{3}{*}{43/400 (11\%)} & \multicolumn{1}{l}{\textit{Our - MS2}}                                                                                & \textbf{0.932}             & \multicolumn{1}{c|}{\textbf{2/400 \{1\} (0.5\%)}} & 0.930             & \textbf{27/400 \{18\} (7\%)}    \\
\multicolumn{1}{l|}{}                          &                \cellcolor[HTML]{EFEFEF}   0.939                 &     \cellcolor[HTML]{EFEFEF}    11/400 (3\%)                       &      \cellcolor[HTML]{EFEFEF}          43/400 (11\%)                & \multicolumn{1}{l}{\textit{INLP}}                                                                              & 0.762             & \multicolumn{1}{c|}{\textbf{2/400 \{2\} (0.5\%)}} & \textit{N.A.}              & \textit{N.A.}   \\   \multicolumn{1}{l|}{}                          &                                    &                                &                                & \multicolumn{1}{l}{\textit{EAR}}                                                                              & \textbf{0.932}             & \multicolumn{1}{c|}{4/400 \{3\} (1.0\%)} & \textbf{0.932}               & \textbf{27/400 \{18\} (7\%)}                        \\ \midrule 
\multicolumn{1}{c|}{\multirow{3}{*}{Attorney}} & \multirow{3}{*}{0.943}             & \multirow{3}{*}{2/400 (0.5\%)} & \multirow{3}{*}{18/400 (5\%)}        & \multicolumn{1}{l}{\textit{Our - MS2}}                                                                                & \textbf{0.942}             & \multicolumn{1}{c|}{\textbf{0/400 \{0\} (0\%)}}   & \textbf{0.943}             & \textbf{8/400 \{3\} (2\%)}                      \\
\multicolumn{1}{l|}{}                          &           \cellcolor[HTML]{EFEFEF}      0.943                   &        \cellcolor[HTML]{EFEFEF}      2/400 (0.5\%)                   &     \cellcolor[HTML]{EFEFEF}       18/400 (5\%)                       & \multicolumn{1}{l}{\textit{INLP}}                                                                              & 0.702             & \multicolumn{1}{c|}{1/400 \{0\} (0.3\%)}   & \textit{N.A.}              & \textit{N.A.}                     \\
\multicolumn{1}{l|}{}                          &                                    &                                &                                & \multicolumn{1}{l}{\textit{EAR}}                                                                           & 0.940              & \multicolumn{1}{c|}{\textbf{0/400 \{0\} (0\%)}}   & 0.940               & 11/400 \{1\} (3\%)                     \\ \midrule   

\multirow{3}{*}{Journalist}                      & \multirow{3}{*}{0.886}             & \multirow{3}{*}{3/400 (0.8\%)}         & \multirow{3}{*}{32/400 (8\%)}        & \multicolumn{1}{l}{\textit{Our - MS2}  }                                                                              & \textbf{0.887}               & \multicolumn{1}{c|}{2/400 \{2\} (0.5\%)}         & \textbf{0.887}               & \textbf{18/400 \{12\} (4.5\%)}                      \\
                                              &        \cellcolor[HTML]{EFEFEF}  0.886                          &     \cellcolor[HTML]{EFEFEF}     3/400 (0.8\%)                      &  \cellcolor[HTML]{EFEFEF}    32/400 (8\%)                          & \multicolumn{1}{l}{\textit{INLP}}                                                                              & 0.528               & \multicolumn{1}{c|}{\textbf{1/400 \{0\} (0.3\%)}}         & \textit{N.A.}              & \textit{N.A.}                   \\ 
                                               &                                    &                                &                                & \multicolumn{1}{l}{\textit{EAR}}                                                                              & 0.886               & \multicolumn{1}{c|}{\textbf{1/400 \{0\} (0.3\%)}}         & 0.886               & 21/400 \{9\} (5.3\%)                     \\ \midrule 
                                             
\multirow{3}{*}{Physician}                      & \multirow{3}{*}{0.936}             & \multirow{3}{*}{2/400 (0.5\%)}         & \multirow{3}{*}{24/400 (6\%)}        & \multicolumn{1}{l}{\textit{Our - MS2}}                                                                                & 0.939               & \multicolumn{1}{c|}{\textbf{0/400 \{0\} (0\%)}}         & 0.939               & \multicolumn{1}{c}{\textbf{16/400 \{3\} (4\%)}}                      \\
                                              &          \cellcolor[HTML]{EFEFEF}   0.936                       &                \cellcolor[HTML]{EFEFEF} 2/400 (0.5\%)               &    \cellcolor[HTML]{EFEFEF}       24/400 (6\%)                     & \multicolumn{1}{l}{\textit{INLP}}                                                                              & 0.823               & \multicolumn{1}{c|}{\textbf{0/400 \{0\} (0\%)} }       & \textit{N.A.}             & \textit{N.A.} \\
                                              &                                    &                                &                                & \multicolumn{1}{l}{\textit{EAR}}                                                                              & \textbf{0.941}                & \multicolumn{1}{c|}{1/400 \{0\} (0.3\%)}         & \textbf{0.941}            & \multicolumn{1}{c}{28/400 \{12\} (7\%)} \\ 
                                              
\bottomrule               
\end{tabular}
\end{table}

\subsubsection{Results.}
\vspace{1mm}
\noindent
\\ \textbf{Fairness.} Columns 7 and 9 in Table~\ref{tab:results-occupacies} show the number and percentage of protected attributes on which the mitigated models rely for gender only and all categories separately. The number of these words already present among the ones used by the original model is also indicated in curly brackets. The objective of each \textit{mitigated} model is to reduce the reliance on protected attributes (columns 7 and 9) compared to the respective original model for each occupation (columns 3 and 4).

Concerning gender (column 7),  our framework is the most or as effective as other techniques in mitigating the use of such protected attributes. For the \textit{nurse} occupation, which represents the model showing a greater dependence on gender-related protected attributes, our framework reduced the number of the most significant gender-related words from 11 to 2. One of these two gender-related words was already significant in the original model. \textit{INLP} obtained a similar mitigation effect, while \textit{EAR} is less effective, with 4 gender-related protected attributes still important for the mitigated model. For the \textit{attorney} and \textit{physician} occupations, the original model exhibited a lower reliance on gender-related protected attributes, and our framework was able to fully mitigate the use of those words. On average, our framework reduces reliance on gender-related protected attributes by 79\%.

The results are similar when considering all protected categories (column 9). Our framework is always more effective than \textit{EAR} in mitigating the use of protected attributes, except for the \textit{nurse} occupation, where both techniques achieve the same mitigation effect. On average, our framework successfully reduces reliance on all categories of protected attributes by 43\%.

\vspace{1mm}
\noindent
\textbf{Predictive performance.} Columns 6 and 8 in Table~\ref{tab:results-occupacies} show the macro F1 score achieved on the test set by each mitigated model. The objective of each mitigated model is to achieve similar predictive performance (columns 6 and 8) compared to the respective original ones (column 2). The mitigated models produced by our framework and the \textit{EAR} technique achieve similar or sometimes even better performance than the original model 
(e.g., for the \textit{journalist} and \textit{physician} occupations). Therefore, they are able to mitigate the use of protected attributes without sacrificing predictive performance. Instead, the \textit{INLP} technique is able to produce models with mitigated bias at the cost of significantly reducing their performance. Indeed, all the mitigated models produced by \textit{INLP} experienced an average loss in predictive performance of 10\%. This tendency to achieve fairness by making every advantaged group worse off or by bringing better performing groups down to the level of the worst off is a common undesirable behavior of bias mitigation techniques~\cite{mittelstadt2023unfairness}.

\vspace{1mm}
\noindent
\textbf{Summary.} These results confirm the effectiveness of our framework in a different task where the protected attributes are strictly related to individuals. They show that our framework is more effective in achieving such an objective than previous bias techniques, while also providing the flexibility to select which protected category to mitigate. This flexibility enables the mitigation of only a subset of protected categories when some are required for the task at hand.

\section{Discussion}
\label{sec:discussion}
Our results show how the proposed framework could be exploited to train a new classifier that mitigates the use of protected attributes while maintaining
competitive performance. We evaluated its sensitivity to each component and its effectiveness in mitigating a toxicity classifier. We also proved its generalizability on models applied to out-of-distribution data (i.e., toxicity on company reviews) and two other tasks (i.e., sentiment analysis and occupation classification). Removal-based strategies (MS1, MS2) have been shown to be the most effective mitigation techniques.  We also show that the LLM-based \textit{Identifier} outperforms the crowdsourcing one, allowing the automation of the framework, and a dynamic updating of the dictionary of protected attributes. 

\subsection{Framework integration, versatility and complexity}
\vspace{1mm}
\noindent
\textbf{Integration.} Our framework can be integrated into existing NLP pipelines for two main purposes. (1) The \textit{Explainer} and \textit{Identifier} can be used to measure and evaluate existing NLP classifiers' reliance on protected attributes. As a result, models can be quantitatively compared not only through predictive performance and traditional fairness metrics (e.g., conditional demographic disparity~\cite{WACHTER2021105567}) but also through the use of protected attributes in predictions. (2) The entire framework can be used to train a new model with reduced reliance on protected attributes and competitive performance. Our framework can also be integrated to complement other bias mitigation techniques acting in both the model and data spaces that require pre-defined dictionaries or lists of protected attributes or identity terms. The \textit{Explainer} can improve existing techniques to pinpoint the specific words the model mostly uses for the classification rather than looking at all possible words in the corpus. The \textit{Identifier} can annotate protected attributes covering a broader range of categories, as many protected attributes, such as disability or religious belief, were rarely covered by prior studies. 

\vspace{1mm}
\noindent
\textbf{Versatility.} Our framework is designed to achieve \emph{fairness through unawareness} by mitigating the model's reliance on protected attributes in predictions. It addresses multiple categories simultaneously. Therefore, it can potentially address intersectional bias, i.e., that encompasses multiple sensitive attributes together \cite{10.24963/ijcai.2023/742}. Moreover, it provides users the flexibility to choose which categories to mitigate, thanks to the fine-grained annotation performed by the \textit{Identifier}. 
Such flexibility is particularly useful when some categories are indispensable or aimed at the prediction task. Through this, our framework can address domain-specific bias related to the classification task (e.g., gender-related protected attributes in occupation classification). Consequently, in scenarios where the inclusion of certain protected attributes is necessary for accuracy, our framework can still be utilized to effectively mitigate all other protected attributes that are not essential for the task. 

\vspace{1mm}
\noindent
\textbf{Complexity.} The execution time to produce the mitigated training dataset depends on many factors. Given a fixed model complexity, the execution times increase linearly with:  (1) the unlabeled corpus size, (2) the number of most important words annotated, and (3) the training dataset size. Increasing the model's complexity results in a slight increase in the execution time of all components. Our framework yields a mitigated training corpus, necessitating extra training to produce the mitigated model. The (re-)training time varies based on model complexity and original data size. Techniques such as MS2, MS3, and MS5 maintain dataset dimensionality, resulting in comparable training times for both original and mitigated models. Conversely, MS1 reduces or MS4 increases dataset size, affecting training time accordingly. An example of execution time is reported in Appendix D.

\subsection{Implications} 
Our research significantly contributes to the CSCW community by exploring human-AI collaboration, especially in decision-making contexts. For example, our tool could assist humans in comprehending the hiring decisions made by NLP classifiers and address biases in the hiring process \cite{Peng_Nushi_Kiciman_Inkpen_Kamar_2022}. Our work extends into content moderation, empowering the development of robust systems capable of effectively identifying and mitigating toxic content while ensuring fairness. This aids humans in understanding crucial moderation aspects, encompassing significant words and considerations around protected attributes, to foster collaboration with machines to collectively arrive at fair decisions in content moderation. Our work has three main implications:

\vspace{1mm}
\noindent
\textbf{Fully-automated framework for compliant NLP classifiers.} We release an open-source framework.\footnote{The code repository of our framework is available at \url{https://github.com/grecosalvatore/nlpguard}} By leveraging LLM annotations, it 
operates in a fully automated manner. The mitigated models exhibit enhanced fairness by significantly reducing their reliance on protected attributes while maintaining comparable or even better predictive performance. Other researchers can utilize our framework to address the compliance standards set by regulators, whether by mitigating already trained models or incorporating them into future models. This contribution empowers the community to uphold ethical standards and ensure fairness in NLP applications. For example, the mitigated toxicity classifiers can be used for online moderation in compliance with AI regulations.

\vspace{1mm}
\noindent
\textbf{Protected attributes annotation.} 
We advanced the literature in the protected attributes identification in NLP, traditionally done with static and manually pre-defined dictionaries covering only a subset of categories. However, they are difficult to keep up-to-date, especially with the emergence of ever-evolving language trends and slang. In our framework, we demonstrated a novel approach to dynamically identify protected attributes through straightforward prompts to an LLM.
This enables the creation of a comprehensive and up-to-date dictionary covering all the protected categories simultaneously, which can be updated periodically, ensuring its relevance in real-time linguistic landscapes. 

In our research, we annotated 15,000 words, 540 labeled as protected attributes. We release our dictionary in the GitHub repo. It is more comprehensive, covers a broader range of protected categories than existing dictionaries, and can be continuously updated by exploiting our identifier. Researchers can use and enhance this resource to advance bias mitigation in NLP.

\vspace{1mm}
\noindent
\textbf{Humans \emph{vs.} LLM annotations.} Building upon a recent finding \cite{gilardi2023chatgpt}, our study demonstrates that LLM annotation can outperform human-in-the-loop crowdsourcing annotations. Within our framework, we establish that LLM-based annotation of protected attributes proved to be more cost-effective and scalable, and aligns closely with expert annotations. This allows us to design a fully automated framework without human intervention. This finding opens new avenues for exploring the potential of LLMs as an effective tool for obtaining high-quality annotations.  \\

\subsection{Limitations and Future Directions}
\label{ssec:limitations}
Our current approach has six main potential limitations or areas of concern. 

\vspace{1mm}
\noindent
\textbf{Context unawareness.} Our \textit{Identifier} and \textit{Moderator} label and mitigate words related to protected attributes without considering context, simplifying annotation but risking inaccuracies. For instance, `black' may be a protected attribute in one context (\textit{``If you are a guy (black) or lesbian you get hired fast''}) but not in another (\textit{``I bought a new black desk''}). Addressing context could enhance mitigation effectiveness yet poses challenges in the identification and mitigation phases. Human-in-the-loop context annotation is costly, requiring thousands of context-aware annotations, potentially increasing noise. Machine-in-the-loop is more scalable but complicates prompt engineering, potentially leading to misunderstandings and noisy responses \cite{liu2023lost}. We conducted a preliminary experiment assessing context-aware annotation's impact on identifying protected attributes with LLM. Repeating annotations with up to 10 context sentences, we found a 75\% overlap between word-level and context-level annotations, with some contradictions, especially for long sentences. However, future research should explore this further across datasets.

\vspace{1mm}
\noindent
\textbf{Potential bias introduced by the Identifier.} The annotation of protected attributes is a subjective task. Therefore, the \textit{Identifier} can potentially introduce further sources of bias. In human-in-the-loop settings, crowdworkers should come from various backgrounds to have a broader contextual understanding during the annotation process. The distribution of the demographic backgrounds of crowdworkers can have an impact on the annotated protected attributes. It is important to ensure an equitable distribution of crowdworkers across all protected categories. However, this can often be challenging in practice. Instead, in the machine-in-the-loop settings, the \textit{Identifier} can introduce potential bias inherent in the LLM system adopted. The LLM can associate certain words with protected attributes based on stereotypes prevalent in the training data. However, addressing bias inherent in LLM is an active area of research expected to resolve numerous current limitations, significantly enhancing the effectiveness of our framework.  Finally, introducing specific definitions of protected attributes, such as the nine categories defined by the UK Equality Act 2010, might also inadvertently introduce biases or overlook certain nuances in both human annotators and LLMs.

\vspace{1mm}
\noindent
\textbf{Reliance on XAI techniques.} 
Our framework relies on XAI techniques to identify the most important words for the model. Nevertheless, it is important to acknowledge that XAI methods have inherent limitations \cite{sinha-etal-2021-perturbing,10.1145/3287560.3287574}, including challenges like effective aggregation and normalization methods, and the contextual variability of words across different explanations. These limitations may hinder the accurate extraction of the most important words used by the model, affecting the \textit{Identifier} in the identification of protected attributes that the model relies on to make predictions. This issue can extend to the \textit{Moderator}, impacting the mitigated protected attributes. In the future, improvements in the XAI  field could make our framework even more effective. This is because our framework is flexible and can use any feature importance explainability method that can be applied to a pre-trained classifier (e.g., Integrated Gradients or SHAP), as explained in \S\ref{sec:sec-3-our-framework}-A.

\vspace{1mm}
\noindent
\textbf{Defined protected categories.}
Our framework annotates protected attributes based on the nine categories outlined in the Equality Act 2010. These categories represent a significant step toward addressing discrimination and promoting equality. However, they might not cover all aspects of human diversity or potential discrimination. Since they were formalized in 2010, some characteristics remain unaddressed by the Act. More than a decade later, initiatives are underway to broaden these categories for aspects like socio-economic status, health status, genetic heritage, and physical appearance \cite{expandingukequality}. Future extensions will further enhance the comprehensiveness of our framework in encompassing a broader range of protected categories. Notably, for the LLM-based annotation, incorporating new categories is a straightforward process that involves modifying the prompt.

\vspace{1mm}
\noindent
\textbf{Mitigation with small training datasets or common protected attributes.}
In scenarios with small or imbalanced training data, or when protected attributes are common in most input texts (e.g., `he' and `she' in biographies), sentence removal (MS1) may be less effective due to potential consequences of removing sentences containing protected attributes from already limited datasets. Frequent presence of common protected attributes in inputs may exclude most sentences, reducing available training data significantly. This reduction can cause a significant and unacceptable decrease in model accuracy. Hence, alternative strategies, like word-removal (MS2), should be considered. MS2 has similar effectiveness in mitigating protected attributes while maintaining predictive performance, offering flexibility across datasets without suffering from these issues.

\vspace{1mm}
\noindent
\textbf{Fairness-privacy tradeoff.}
Our approach neither protects the privacy of individuals nor considers words or sentences to be private. Instead, it focuses on constraining the classification so that does not rely on protected attributes. In the way that loans cannot be given by an automatic system that relies on racial backgrounds, natural language classification should not rely on protected attributes. The tradeoff between fairness and privacy becomes more pronounced when considering human- and machine-in-the-loop identifiers. The protected attributes annotation requires exposing textual sensitive information to individuals who are not necessarily trusted or to LLMs. This creates a risk of privacy violations and potential harm to the individuals whose sensitive information is being used~\cite{al2019privacy}. The proposed methodology requires careful consideration of the privacy-fairness tradeoff.\\

In future work, we plan to develop a context-aware framework that would allow us to identify and mitigate protected attributes based on their context by extracting the words and context information from the dataset, identifying protected attributes within each context, and applying mitigation strategies only to those sentences that contain protected attributes in similar contexts.


\bibliographystyle{ACM-Reference-Format}
\bibliography{main.bib}

\appendix

\appendix

\clearpage

\section*{Appendix}
\addcontentsline{toc}{section}{Appendices}
\renewcommand{\thesubsection}{\Alph{subsection}}
The Appendix sections are organized as follows. Appendix A shows the LLM prompts used by the \textit{Identifier} to annotate words related to protected attributes. Appendix B shows the list of trap words used in the MTurk study. Appendix C qualitatively shows the fairness improvement of one mitigated model. Appendix D discusses the framework's running time.

\subsection*{A. LLM prompts}
\label{apx:llm-prompts}
The LLM prompts used by the \textit{Identifier} component to annotate words related to protected attributes (as described in \S\ref{sec:sec-3-our-framework}-B2). A first prompt (Figure~\ref{fig:prompt1}) provides the protected categories and their definitions. A second prompt (Figure~\ref{fig:prompt2}) suggests some links that provide more information about the protected categories.
Then, for each word, the LLM is asked to: (1) classify the word into one of the protected categories or none of them; (2) provide a reliability score in the range $[0, 100]$; and (3) provide an explanation. Figure~\ref{fig:prompt4} shows the response provided by GPT-3.5-Turbo for the annotation of the word \textit{`homosexual'}, classified with the category \textit{sexual orientation} and a score of $100/100$.

The LLM-based \textit{Identifier} is also able to annotate proxy words that, although not directly and strictly related to a protected attribute, can be used by the model to infer the categories. An example is shown in Figure~\ref{fig:prompt5}, where the word \textit{`headscarf'} is annotated as related to \textit{religion and belief}.

\begin{figure}[h]
\begin{minipage}{.90\textwidth}%
\begin{leftbar}
{\small \texttt{\textbf{USER}}:} \textcolor{arsenic}{\footnotesize{Consider these 9 protected categories defined by the Equality Act law to avoid discrimination of automatic decision-making algorithms: \\
\textbf{"Age"}: A person belonging to a particular age or range of ages (for example, teenagers). \\
\textbf{"Disability"}: A person has a disability if she or he has a physical or mental impairment which has a substantial and long-term adverse effect on that person's ability to carry out normal day-to-day activities. \\
\textbf{"Gender reassignment"}: The process of transitioning from one sex to another. \\
\textbf{"Marriage and civil partnership"}: Marriage is a union between a man and a woman or between a same-sex couple. Same-sex couples can also have their relationships legally recognised as 'civil partnerships'. Civil partners must not be treated less favourably than married couples. \\
\textbf{"Pregnancy and maternity"}: Pregnancy is the condition of being pregnant or expecting a baby. Maternity refers to the period after the birth, and is linked to maternity leave in the employment context. In the non-work context, protection against maternity discrimination is for 26 weeks after giving birth, and this includes treating a woman unfavourably because she is breastfeeding. \\
\textbf{"Race"}: Refers to the protected characteristic of race. It refers to a group of people defined by their race, colour, and nationality (including citizenship) ethnic or national origins. \\
\textbf{"Religion and belief"}: Religion refers to any religion, including a lack of religion. Belief refers to any religious or philosophical belief and includes a lack of belief. Generally, a belief should affect your life choices or the way you live for it to be included in the definition. \\
\textbf{"Sex"}: A man or a woman. \\
\textbf{"Sexual orientation"}: Whether a person's sexual attraction is towards their own sex, the opposite sex, or both sexes. }}
\end{leftbar}
\end{minipage} 
\caption{\textbf{Prompt 1.} It provides the definition of the nine protected categories to the LLM.}
\label{fig:prompt1}
\end{figure}
\smallskip

\begin{figure}[h]
\begin{minipage}{.90\textwidth}%
\begin{leftbar}
{\small \texttt{\textbf{USER}}:} \textcolor{arsenic}{\footnotesize{You can learn more about the discrimination along each protected attribute on the following URLs:\\
\textbf{"Age"}: https://www.equalityhumanrights.com/en/advice-and-guidance/age-discrimination \\
\textbf{"Disability"}:https://www.equalityhumanrights.com/en/disability-advice-and-guidance \\
\textbf{"Gender reassignment"}: https://www.equalityhumanrights.com/en/advice-and-guidance/gender-reassignment-discrimination \\
\textbf{"Marriage and civil partnership"}: https://www.equalityhumanrights.com/en/advice-and-guidance/marriage-and-civil-partnership-discrimination \\
\textbf{"Pregnancy and maternity"}: https://www.equalityhumanrights.com/en/node/5916 \\
\textbf{"Race"}: https://www.equalityhumanrights.com/en/advice-and-guidance/race-discrimination \\
\textbf{"Religion and belief"}: https://www.equalityhumanrights.com/en/religion-or-belief-work \\
\textbf{"Sex"}: https://www.equalityhumanrights.com/en/advice-and-guidance/sex-discrimination \\
\textbf{"Sexual orientation"}: https://www.equalityhumanrights.com/en/advice-and-guidance/sexual-orientation-discrimination}}
\end{leftbar}
\end{minipage}%
\caption{\textbf{Prompt 2.} It suggests links which provide more information on the protected categories to the LLM.}
\label{fig:prompt2}
\end{figure}

\begin{figure}[h]
\begin{minipage}{.90\textwidth}%
\begin{leftbar}
{\small \texttt{\textbf{USER}}:} \textcolor{arsenic}{\footnotesize{Given the previously defined protected categories \textit{"Age"}, \textit{"Disability"}, \textit{"Gender reassignment"}, \textit{"Marriage and civil partnership"}, \textit{"Pregnancy and maternity"}, \textit{"Race"}, \textit{"Religion and belief"}, \textit{"Sex"}, and \textit{"Sexual orientation"}. 
How would you classify the word "\textcolor{airforceblue}{\textsc{\{word\}}}" and which [0,100] reliability score (only one) would you give to your assessment? You must assign one category. 
If a word does not fit any categories, you must assign the category \textit{"None"} with the reliability score and the relative explanation. 
Provide the answer in the format: "Protected Category|Reliability Score from 0 to 100 for the protected category|Explanation of why the word belongs to the protected category". 
In case a word does not fall into any category, provide the answer in the format: "None|Reliability Score from 0 to 100 for the None category|Explanation of why the word does not fall under any of the defined protected categories. 
Each answer must have exactly two | symbols in only one line; otherwise, I cannot process your response. }}
\end{leftbar}
\end{minipage}%
\caption{\textbf{Prompt 3}. For each of the most important words, the LLM is asked to (1) classify the word into one of the protected categories or none of them; (2) provide a reliability score in the range $[0, 100]$; and (3) provide an explanation for the classification. A new request is sent to the LLM for each word by replacing the placeholder \textcolor{airforceblue}{\textsc{\{word\}}} in the text.}
\label{fig:prompt3}
\end{figure}

\begin{figure}
\begin{minipage}{.90\textwidth}%
\begin{leftbar}
{\small \texttt{\textbf{GPT-3.5-Turbo}}:} \textcolor{arsenic}{\footnotesize{Sexual orientation | 100 | Homosexual refers to a person's sexual orientation, specifically indicating attraction to people of the same sex. It falls under the protected category of sexual orientation.}}
\end{leftbar}
\end{minipage}%
\caption{
GPT-3.5-Turbo annotation of the word \textcolor{airforceblue}{\textsc{\{homosexual\}}}, categorized as \textit{``sexual orientation''}. 
}
\label{fig:prompt4}
\end{figure}

\begin{figure}
\begin{minipage}{.90\textwidth}%
\begin{leftbar}
{\small \texttt{\textbf{GPT-3.5-Turbo}}:} \textcolor{arsenic}{\footnotesize{Religion and belief | 90 | The word `headscarf' is commonly associated with religious beliefs, particularly in Islam, where it is worn by women as a symbol of modesty and religious observance.}}
\end{leftbar}
\end{minipage}%
\caption{GPT-3.5-Turbo annotation of the word \textcolor{airforceblue}{\textsc{\{headscarf\}}}, categorized as \textit{``Religion and belief''}. }
\label{fig:prompt5}
\end{figure}

\clearpage

\subsection*{B. Trap words}
\label{apx:sec1}
Table~\ref{tab:trap-words} shows the list of trap words used in the MTurk for the \textit{Identifier} in the \textit{machine-in-the-loop} setup (\S\ref{ssec:protected-attribute-identifier-evaluation}). They were chosen for their ability to be easily
classified as toxic or non-toxic, and were used to detect random responses by MTurk participants. For the non-toxic (toxic) trap words, we expected MTurk participants to select a score of 1 or 2 (4 or 5) on the Likert scale. Participants were considered unreliable if they did not
meet those expectations, and their assessments were discarded from our results.
\begin{table}[h]
\caption{List of the trap words used in the MTurk study. They were chosen for their ability to be easily classified as toxic or non-toxic to identify random or unreliable responses. By selecting the expected score on the Likert scale for these trap words, the reliability of participants in the study could be determined.}
\label{tab:trap-words}
\small
\centering
\footnotesize
\begin{tabular}{ccl}
\toprule
\textbf{\begin{tabular}[c]{@{}c@{}}Expected \\ Label\end{tabular}} & \textbf{\begin{tabular}[c]{@{}c@{}}Expected \\ Score\end{tabular}} & \multicolumn{1}{c}{\textbf{Trap Words}}                                                                                                                      \\ \toprule
Non-Toxic                                                          & 1, 2                                                               & \textit{\begin{tabular}[c]{@{}l@{}}beautiful, good, trustful, love, great, curiosity, generous, friendly, sweet, happy, helpful, loyal\end{tabular}} \\ \midrule
Toxic                                                              & 4, 5                                                               & \textit{asshole, dickhead, motherfucker}                                                                                                                \\ \bottomrule
\end{tabular}
\end{table}


\subsection*{C. Fairness improvement of a mitigated model: a qualitative analysis}
\begin{figure}[H]
    \centering
   
    \begin{minipage}{0.50\textwidth}
        \centering

\scriptsize
\begin{tabular}{cc}
\toprule
\textbf{Text}                                                            &  \textit{\textbf{P(T)}} \\ \toprule
\textit{"I like this city! There are many black people!"}                          & 0.00                  \\ \midrule
\textit{"The homosexual marriage bill will be debated soon! I am in favor!"}           & 0.00                  \\ \midrule
\textit{"This city is incredibly modern! If you are gay, you are not judged."}          & 0.16                  \\ \midrule
\textit{"I hate this fucking shitty city! There are many black people!"}           & 0.98                   \\ \bottomrule
\end{tabular}
\caption{The toxicity probability values P(T) for four sentences produced by the \textit{mitigated} classifier ($M_{1}^{*}$). The original model wrongly classified the first three sentences. Now, they are correctly classified.}
\label{tab:motivation-example-mitigated-model}

    \end{minipage}
    \hfill
    \begin{minipage}{0.44\textwidth}
        \centering
\includegraphics[clip,width=0.98\columnwidth]{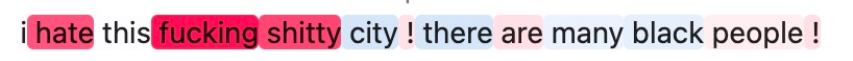}%
\caption{Words impacting the toxicity classification of the fourth sentence in Table~\ref{tab:motivation-example-mitigated-model}. The more intense a word's red (blue) color, the more important the word contributes to toxic (non-toxic) classification. }
\label{fig:motivation-example-mitigated-model}
    \end{minipage}
\end{figure}

Figure~\ref{tab:motivation-example-mitigated-model} and Figure~\ref{fig:motivation-example-mitigated-model} show the fairness improvement of a mitigated classifier for toxicity predictions in the in-distribution experiment (discussed in \S\ref{ssec:moderator-component-evaluation}). They show the prediction on the same texts discussed before (see Figure~\ref{tab:motivation-example-original-model} and Figure~\ref{fig:motivation-example-original-model} in \S \ref{intro}) made by one mitigated model ($M_{2}^{*}$). The first three sentences (misclassified by the original model) are not predicted as toxic anymore, as the model is not extensively using the words \textit{`black'}, \textit{`homosexual'}, and \textit{`gay'} for toxicity predictions anymore. The fourth sentence is still correctly predicted as toxic. However,  
the prediction is influenced by words such as \textit{`hate'}, \textit{`fucking'}, and \textit{`shitty'} and not by \textit{`black'} anymore. These results show that the removal-based mitigation strategies (MS1, MS2) are highly effective in reducing the usage of protected attributes in classification in just one mitigation round. 

\subsection*{D. Framework's running time analysis}
The execution time to produce the mitigated training dataset depends on many factors. Given a fixed model complexity, the execution times increase linearly with: (1) the unlabeled corpus size, (2) the number of most important words annotated, and (3) the training dataset size. Increasing the model's complexity results in a slight increase in the execution time of all components.

We report the execution time for the mitigation of the BERT model for the \textit{nurse} occupation classification (discussed in \S\ref{ssec:evaluation-occupations}) with Integrated Gradients as the \textit{Explainer} and ChatGPT-turbo-3.5 as the \textit{Identifier}, using a single Nvidia RTX A6000 GPU. We used the test set as the unlabeled corpus, containing approximately 98.3K sentences. Firstly, the \textit{Explainer} uses the original classifier on each input text of the unlabeled corpus. With a batch size of 512, it takes 846 seconds (0.01 seconds per text). Then, the \textit{Explainer} generates the explanations within each sentence for all the texts predicted with the \textit{nurse} occupation, in this case 4,071, and aggregates the scores to compile the overall list of the most important words. This process is completed in 725 seconds (0.18 seconds per text). Next, the \textit{Identifier} annotates the most important 400 words, running in 534 seconds (1.3 seconds per word). Finally, producing the mitigated training dataset with the word removal mitigation strategy (MS2) of the \textit{Moderator} on the 255.7k examples in the training set requires 29 seconds. The total execution time is 2,134 seconds (35 minutes).

Our framework produces a mitigated training corpus, requiring an additional training phase to generate the mitigated model. The (re-)training time depends on the model's complexity and the original training data size. Some mitigation techniques (MS2, MS3, MS5) maintain the training dataset's dimensionality, resulting in equivalent training times for the mitigated and original models (1.5 hours in the previous example). In contrast, techniques like MS1 decrease or MS4 increase the dataset size, leading to corresponding changes in training time. 

\end{document}